\documentclass[twoside,twocolumn,9pt]{article}
\usepackage{extsizes}
\usepackage[super,sort&compress,comma]{natbib} 

\usepackage[left=1.5cm, right=1.5cm, top=1.785cm, bottom=2.0cm]{geometry}
\usepackage{balance}

\usepackage{verbatim}
\usepackage{mathtools}
\usepackage{subcaption}
\usepackage{graphicx}
\usepackage{bm}
\usepackage{makecell}
\usepackage{multirow}

\usepackage{mathptmx}
\usepackage{sectsty}
\usepackage{graphicx} 
\usepackage{lastpage}
\usepackage[format=plain,justification=justified,singlelinecheck=false,font={stretch=1.125,small,sf},labelfont=bf,labelsep=space]{caption}
\usepackage{float}
\usepackage{fnpos}
\usepackage[english]{babel}
\addto{\captionsenglish}{%
  \renewcommand{\refname}{Notes and references}
}
\usepackage{array}
\usepackage{droidsans}
\usepackage{charter}
\usepackage[T1]{fontenc}
\usepackage[usenames,dvipsnames]{xcolor}
\usepackage{setspace}
\usepackage[compact]{titlesec}
\usepackage{hyperref}
\usepackage{cleveref}

\usepackage{epstopdf}

\definecolor{cream}{RGB}{222,217,201}

\begin{document}

\thispagestyle{plain}


\makeFNbottom
\makeatletter
\renewcommand\LARGE{\@setfontsize\LARGE{15pt}{17}}
\renewcommand\Large{\@setfontsize\Large{12pt}{14}}
\renewcommand\large{\@setfontsize\large{10pt}{12}}
\renewcommand\footnotesize{\@setfontsize\footnotesize{7pt}{10}}
\makeatother

\renewcommand{\thefootnote}{\fnsymbol{footnote}}
\renewcommand\footnoterule{\vspace*{1pt}%
\color{cream}\hrule width 3.5in height 0.4pt \color{black}\vspace*{5pt}} 
\setcounter{secnumdepth}{5}

\makeatletter 
\renewcommand\@biblabel[1]{#1}            
\renewcommand\@makefntext[1]%
{\noindent\makebox[0pt][r]{\@thefnmark\,}#1}
\makeatother 
\renewcommand{\figurename}{\small{Fig.}~}
\sectionfont{\sffamily\Large}
\subsectionfont{\normalsize}
\subsubsectionfont{\bf}
\setstretch{1.125} 
\setlength{\skip\footins}{0.8cm}
\setlength{\footnotesep}{0.25cm}
\setlength{\jot}{10pt}

\makeatletter 
\newlength{\figrulesep} 
\setlength{\figrulesep}{0.5\textfloatsep} 

\newcommand{\topfigrule}{\vspace*{-1pt}%
\noindent{\color{cream}\rule[-\figrulesep]{\columnwidth}{1.5pt}} }

\newcommand{\botfigrule}{\vspace*{-2pt}%
\noindent{\color{cream}\rule[\figrulesep]{\columnwidth}{1.5pt}} }

\newcommand{\dblfigrule}{\vspace*{-1pt}%
\noindent{\color{cream}\rule[-\figrulesep]{\textwidth}{1.5pt}} }

\makeatother


\sffamily
\title{Random projections and Kernelised Leave One Cluster Out Cross-Validation: Universal baselines and evaluation tools for supervised machine learning for materials properties} 

 \author{Samantha~Durdy,$^{\ast}$\textit{$^{abc}$} Michael~Gaultois,\textit{$^{bcd}$}, Vladimir~Gusev,\textit{$^{bcd}$} Danushka~Bollegala,\textit{$^{ac}$} and Matthew~J.~Rosseinsky\textit{$^{bcd}$}} 
 \date{June 2022}
\maketitle

\abstract{With machine learning being a popular topic in current computational materials science literature, creating representations for compounds has become common place. These representations are rarely compared, as evaluating their performance -- and the performance of the algorithms that they are used with -- is non-trivial. With many materials datasets containing bias and skew caused by the research process, leave one cluster out cross validation (LOCO-CV) has been introduced as a way of measuring the performance of an algorithm in predicting previously unseen groups of materials. This raises the question of the impact, and control, of the range of cluster sizes on the LOCO-CV measurement outcomes. We present a thorough comparison between composition-based representations, and investigate how kernel approximation functions can be used to better separate data to enhance LOCO-CV applications. 
\par We find that domain knowledge does not improve machine learning performance in most tasks tested, with band gap prediction being the notable exception. We also find that the radial basis function improves the linear separability of chemical datasets in all 10 datasets tested and provide a framework for the application of this function in the LOCO-CV process to improve the outcome of LOCO-CV measurements regardless of machine learning algorithm, choice of metric, and choice of compound representation. We recommend kernelised LOCO-CV as a training paradigm for those looking to measure the extrapolatory power of an algorithm on materials data.} \\


\renewcommand*\rmdefault{bch}\normalfont\upshape
\rmfamily
\section*{}
\vspace{-1cm}


\footnotetext{\textit{$^{a}$~Department of Computer Science, University of Liverpool, Ashton Street, Liverpool, L69~3BX, UK}}
\footnotetext{\textit{$^{b}$~Materials Innovation Factory,
University of Liverpool, 51 Oxford Street, Liverpool, L7~3NY, UK}}
\footnotetext{\textit{$^{c}$~Leverhulme Research Centre for Functional Materials Design, University of Liverpool, 
Oxford Street, Liverpool, L7~3NY, UK}}
\footnotetext{\textit{$^{d}$~Department of Chemistry, University of Liverpool, Crown St, Liverpool, L69~7ZD, UK}}

\footnotetext{\textit{$\ast$E-mail: samantha.durdy@liverpool.ac.uk}}




\section{Introduction}
Recent advances in materials science have seen a plethora of research into application of machine learning (ML) algorithms. Much of this research has focused on supervised ML methods, such as random forests (RFs) and neural networks. More recently, authors have laid out the best practices to help unify and progress this field~\cite{SchmidtReview,WardReview,butlerReview,BestPractices}.
\par Data representation can play a large role in the performance of ML algorithms; however, optimum choice of representation is not always apparent. In materials science it is often difficult to choose an appropriate representation due to variability in the ML task and in the nature of the chemistry, composition and structures of the materials studied. Additionally, some properties of a material, such its crystal structure in the case of crystalline materials, may not been known until its synthesis. Accordingly, many studies derive representations from either the ratios of elements in the chemical composition, or from domain knowledge- based properties (referred to as features) of these elements, or both, in a process called “featurisation.”.  
\par Given the ubiquity of featurisation methods such as those presented here in materials applications, it is important to evaluate the statistical advantage of specific feature sets~\cite{Murdock}. Section~\ref{representations} overviews different featurisation techniques and how their effectiveness has been previously reported. We expand on this evaluation in~\cref{caseStudies}, in which seven representations are investigated across five case studies from the literature to explore how these representations perform in published ML tasks. These cases thus represent practical applications, rather than constructed tasks. Each of these representations is also compared to a random projection of equal size to establish the performance benefit of domain knowledge over random noise.
\par Evaluating the generalisability of ML models is a known challenge across data science, and is of particular concern in materials science, where data sets are of limited size compared with other application areas for ML, and often biased towards historically interesting materials or those closely related to known high-performance materials for certain performance metrics. Typically, models are evaluated on test sets separate from their training data, through a consistent train-test split or N-fold cross validation. However, this does not consider skew in a dataset. In chemical datasets, families of promising materials are often explored more thoroughly than the domain as a whole, which introduces bias and reduces the generalisability of ML models because the data they are trained and tested on are not sampled in a way representative of the domain of target chemistries to be screened with these models.
\par Leave one cluster out cross validation (LOCO-CV) was suggested to combat this~\cite{LOCOCV}, using K-means clustering to exclude similar families of materials from the training set to measure the extrapolatory power of an ML algorithm (its ability to predict the performance of materials with chemistries qualitatively different from the training set). The value of such an approach can be seen in the case of predicting new classes of superconductors. One may choose to remove cuprate superconductors from the training set, and if an ML model can then successfully predict the existence of cuprate superconductors without prior knowledge of them, we can conclude that that model is likely to perform better at predicting new classes of superconductors than a model which could not predict the existence of cuprate superconductors. LOCO-CV provides an algorithmic framework to measure the performance of models on predicting new classes of materials by defining these classes as clusters found by the K-means clustering algorithm. Application and implementation of this algorithm is discussed further in~\cref{lococv}.
\par While differences in cluster sizes in this domain are expected, it has been observed that clusters found with K-means can differ in size by orders of magnitude~\cite{JonsPaper}, which can pose a practical challenge to adoption of this method. With such differences in cluster size, LOCO-CV measurements can represent the performance of an algorithm on a small training set rather than the performance of an algorithm in extrapolation. As representation plays a role in clustering, it is pertinent to investigate the issues of representation and clustering together, even though the representation used in clustering does not need to be the same as that used to train the model (\cref{fig:flowChart})
In~\cref{linearSeperability} we investigate how representations can affect measurements made with LOCO-CV. Kernel methods (also known as kernel tricks, or kernel approximation methods), can be used to non-linearly translate data into a data space that can then be linearly separated (\cref{fig:dummy_rbf}). We apply kernel methods such as the radial basis function (RBF) to chemical datasets to improve the linear separability of data and reduce variance between cluster sizes and thus increase the validity of LOCO-CV measurements (\cref{fig:ICSDPerformance,fig:LOCOCVConvergence}), thus enhancing the assessment of performance found when using different representations as well as assessment of model performance as a whole.
\par LOCO-CV evaluation is affected by representation of a compound and, conversely, choice of compound representation is affected by the methods used to evaluate these representations. Thus, it is pertinent to investigate these two issues simultaneously. We improve the utility of LOCO-CV measurements by using kernel functions to create a more separable data space, and use these measurements to evaluate featurisation methods using practical supervised ML tasks found in the literature. The key contributions of this paper are as follows:
\begin{itemize}
    \item Further comparison between composition-based feature vectors by comparing performance measured when using different featurisation methods on practical tasks (explained further in~\cref{representations} before being carried out in section ~\cref{caseStudies}).
    \item Examining the effectiveness of random projections as featurisation methods, as a baseline to justify more involved featurisation methods against (explained further in \cref{randomVectors} before being carried out in~\cref{caseStudies}).
    \item Novel application kernel methods to materials data (explained further in \cref{kernelMethods} before being carried out in~\cref{linearSeperability}).
    \item Studying the effect of kernel approximation functions on application of K-means clustering to materials data and presenting a workflow to incorporate these methods into the LOCO-CV algorithm (\cref{linearSeperability}).
    \item We recommend using RBF when clustering for LOCO-CV, as clusterings found after application of RBF are seen to be more even in size than with no kernel method applied, and to give more reliable model convergence. This helps to reduce the risk that performance differences on predicting an unseen cluster of data are caused by the training set size as opposed to the intrinsic inability of a model to perform well on that cluster of data. 
    \item We note that use of the radial basis function (RBF) in clustering for LOCO-CV results in models converging (\textit{i.e.} learning trends from the data to be able to make predictions with some degree of reliability) more often than when using LOCO-CV without any kernel methods.
    \item We suggest that random projections are used as a baseline against which to compare engineered feature vectors, noting that commonly used CBFVs have little to no advantage over random projections in most tasks tested here.
    \item We experiment with the use of random projections in clustering for LOCO-CV, finding them to have no clear advantage over other CBFVs tested in this task.
\end{itemize}

\section{Key Concepts and Techniques}
\begin{figure*}
    \centering
    \begin{subfigure}{0.75\linewidth}
    \centering
    \includegraphics[width=\linewidth]{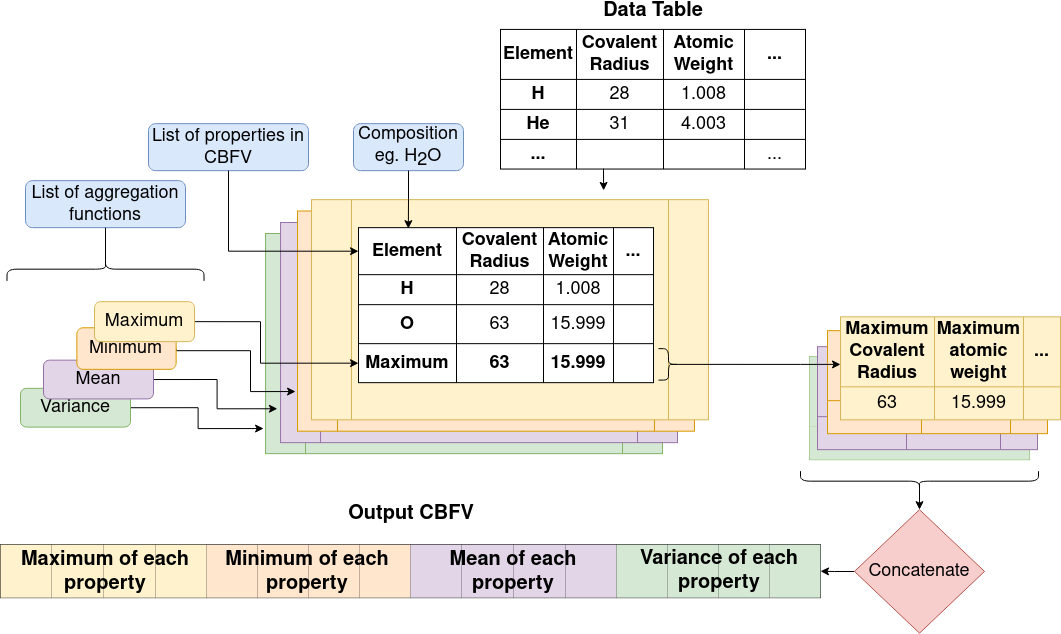}
    \caption{}
    \label{fig:CBFVworkflow}
\end{subfigure}
    \centering
    \begin{subfigure}{0.75\linewidth}
    \centering
    \includegraphics[width=\linewidth]{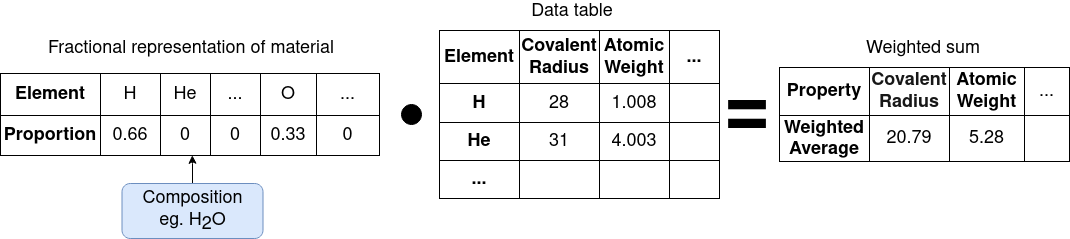}
    \caption{}
    \label{fig:weightedSum}
    \end{subfigure}
    \hfill
    \begin{subfigure}{0.75\linewidth}
    \centering
    \includegraphics[width=\linewidth]{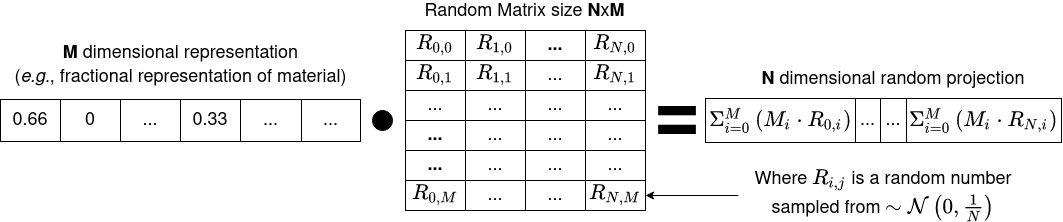}
    \caption{}
    \label{fig:RandomProjection}
    \end{subfigure}
    \caption[Comparison of the creation composition based feature vectors and random projections.]{Comparison of the creation of composition based feature vectors (CBFVs) and random projections. (a) General workflow for creation of CBFV. Application of aggregation function to each property of a material will result in a fixed sized vector for each aggregation function, these are then concatenated together (merged sequentially) to form the final CBFV. Both the properties in the CBFV and the list of aggregation functions can be changed to create variants of CBFVs, which may influence algorithms that use the resulting CBFV. (b) Calculation of the weighted sum of properties of a material. This is equivalent to the matrix multiplication of the fractional representation of that material and its properties. (c) Calculation of a random projection. Using random projection to (approximately) linearly project a representation into a different number of dimensions ($N$). The original $M$ dimensional representation for our purposes may be a fractional representation for the chemical composition of a material, but this technique can be used for any input data, in domains outside of chemistry.}
    \label{fig:matrixMultiplication}
\end{figure*}
\subsection{Common representations used in machine learning in inorganic chemistry}~\label{representations}
\par ML algorithms require a consistent definition of a data point in order to analyse trends within a dataset. For example, it would be hard to learn from a dataset in which ``a data point'' may refer to a phase field, a specific crystal structure, or a composition. One such algorithm is RFs, which are widely used in materials science as well as other domains\cite{breiman}. They are fast to train, readily implemented \cite{skl}, and see a good performance in a plethora of tasks without hyperparameter tuning. We use RFs for our investigations for reasons outlined above, however good evaluation methods for fixed dimensional representations of materials are also important for the plethora of other ML algorithms that use such representations as basis for predictions.
\par Representation learning, and feature engineering are the two main preprocessing methods to make data more interpretable to ML algorithms. Representation learning is a fast-evolving field that uses deep learning in order to create representations, while feature engineering involves defining a set of features (or descriptors) for a data point that adequately encapsulates all information needed~\cite{Bengio2013}.
\par Feature engineering has been used extensively in inorganic chemistry and materials science. However, no set of features has emerged as the clearly dominant representation for a material, likely due to the variety of tasks carried out in these domains, which may require different input representation. Many of these representations use only composition-based information (rather than structural), as this allows screening of materials without need for DFT calculations or synthesis, greatly reducing costs associated with such screenings. Composition-based screening is less powerful than the incorporation of structure, as both structure and composition control properties, but more general as structural information is not required and is less widely available than composition (as structure is not known until the material is realised by synthesis, whereas compositions can be proposed without knowing structure). Composition-based feature vectors (CBFVs), which offer a list of compositional attributes of a material, and a one-hot style (also called fractional) encoding of composition~\cite{elemnet}, are widely used composition-based representations.
\par Notable CBFVs including $Magpie$, $Oliynyk$ and $JARVIS$~\cite{Ward2016,Oliynyk,JARVIS} (differences between which are discussed further during~\cref{caseStudies}) were recently investigated and found to provide benefit over one-hot style representations. This benefit was measured using neural networks predicting numerous properties, however the benefit became little to none as the dataset size increased above 1000 points~\cite{Murdock}. 
\par We further the investigation into the use of CBFVs by examining their applicability in five case studies. Namely, we examine performance using \textit{Olinyk}, \textit{Magpie}, and \textit{JARVIS}, a variant of random projection of size 200 (discussed more in~\cref{randomVectors}) used in the previous review on this topic~\cite{Murdock}, as well as one-hot style encodings of composition, and random linear projection of the composition. The performance of RFs using different representations are compared on ML tasks found in the literature, using the relevant datasets for each study~\cite{Stanev,Legrain2018,Ward2018,Davies2019,Sparks2018}. 
\par The representations were chosen as they are commonly used, and as these are the non-structural representations investigated for their efficacy in neural networks in previous work~\cite{Murdock}. Seeing whether previous results hold for RFs should help gauge whether these results could be used as rule of thumb for many ML algorithms or whether these conclusions should only be applied to neural networks similar to those used in that study.

\subsubsection{Can implementation details in CBFVs affect performance}~\label{implementationDetails}
\par It is common for a CBFV to be comprised of a list of elemental properties that are combined using several ``aggregation functions'', for example the weighted average, and standard deviation of various elemental properties in a compound (\cref{fig:CBFVworkflow}). The aggregation functions of a CBFV can vary between implementations~\cite{Ward2016,Murdock}. Using different numbers of aggregation functions results in representations of different lengths (\cref{fig:CBFVworkflow}), which may affect ML performance depending on the algorithm being used.
\par Problems associated with building statistical models using increasingly large data representations without also increasing the number of data points are well documented, often being described as the curse of dimensionality~\cite{curseOfDimensionality}. Strong correlation between different dimensions (known as co-linearity, or cross correlation between dimensions) can also impact model performance. For example, RFs are affected by co-linearity between dimensions as RF’s random bagging process is unlikely to select a subset of features that include none of a set of cross corelated features. This would make the information in features with such cross-corelates more likely to be available to discriminate with at any branch in a tree, compared with those features without such cross-corelates. It is intuitive that different aggregation functions may be cross-correlated, for example the maximum atomic weight of an element in a compound is likely to correlate with the average atomic weight of an element in that compound, thus RFs may be affected by additional aggregation functions.
\par Without investigation, it is unclear what effect different aggregation functions will have on algorithm performance. Interrogation of the repository associated with the previous review of featurisation methods indicates use of the weighted average, sum, range, and variance of each feature~\cite{Murdock}. This includes the features of the fractional (one-hot style) representation, which uses only the ratios of each element in a material in its definition. This implementation difference could affect the performance of a model that uses these representations, so we distinguish between the two, using ``fractional'' to refer to a one-hot style encoding that includes the average, sum, range, and variance of each element and ``CompVec'' (for composition vector) to refer to an implementation of one-hot style encoding which contains just the ratios of elements in a compound. 
\par The nature of the fractional representation means that a given compound would contain the same representation three times, scaled by different amounts (depending on the number of elements in the compound) in a single vector (four times if elements in a compound are in equal ratios). This can be exemplified by examining a simple composition such as NaCl (\cref{tab:nacl}).
.
\begin{table}
    \centering
    \begin{tabular}{c|c|c|c}
         Aggregation function & Na & Cl & All other columns \\\hline
         weighted average & 0.5 & 0.5 & 0 \\
         sum & 1 & 1 & 0\\
         range & 1 & 1 &0\\ 
         variance & 0.0042 & 0.0042 &0
         
    \end{tabular}
    \caption{Values that would occur in each column across different aggregation functions for a composition fractional representation of NaCl. This demonstrates how the inclusion of additional aggregation functions does not add additional information for this representation. These calculations assume a representation which allows for 118 different elements, a smaller number of represented elements would result in the values in the variance columns being larger.}
    \label{tab:nacl}
\end{table}
\par This offers an opportunity to investigate how increasing dimensionality (the number of dimensions) of a representation while adding no new information affects performance. We leave the investigation of the effect of information added by different aggregation functions on different feature sets to future work. We experiment using both a (\textit{CompVec}) one-hot style encoding as proposed for use with ElemNet\cite{elemnet} (with no additional aggregation functions), and the one-hot style approach used previously that includes different aggregation functions (\textit{fractional})~\cite{Murdock}, to see how this increase in dimensionality above will affect experiments.

\par While this increase in dimensionality will be seen to affect the clusterings found with K-means clusterings, for most tasks investigated there was not an appreciable difference between CompVec and fractional representations. In band gap prediction tasks fractional representation outperformed CompVec, however in regression tasks relating to bulk metallic glass formation this trend was reversed (\cref{fig:CaseStudies}).

\subsubsection{Random Vectors as featurisation methods}\label{randomVectors}
Each elemental property (for example covalent radius) aims to bring with it some sort of information about that element. That property’s inclusion in a feature set aims to improve an ML algorithm’s performance in a given problem. Every feature included either means an increase to the dimensionality of a CBFV or the exclusion of an alternative feature. Though the importance of a feature to an ML model can be measured~\cite{GiniFI,permutationFI}, it is hard to take such measures of feature importance out of the context of the model that is trained with it, or the dataset that the model is derived from~\cite{univariateFeatureSelection}.
\par As it is hard to distinguish the effects of dimensionality of a representation from the effects of the information imbued in it, Murdock \textit{et al.} introduce a set of vectors, one for each element each consisting of 200 random numbers to represent nonsensical elemental properties. From these vectors, they derive the CBFV \textit{RANDOM\_200} to represent a lower bound for feature performance. That is to say; rather than using features that would be expected to give information about an element (covalent radius, atomic number etc.), they instead assign each element a vector of random numbers. If these random numbers can result in a well-performing model then whether the chemically-derived features that are commonplace in the literature are justified can be called into question. When the aggregation function is a weighted sum (discussed further in \cref{implementationDetails}), this has the same effect as a matrix multiplication of the one-hot style encoding of a compounds formulae, $\bm{C}$, (referred to in this paper as \textit{CompVec}), and a random matrix, $\bm{R}$~which can be noted as $\bm{C\cdot R}$ (\cref{fig:weightedSum}). Thus the weighted sum part of the \textit{RANDOM\_200} can be seen as a matrix multiplication of the random vectors and the fractional encoding of the composition.

\par This matrix multiplication is similar to that used in a random projection. Random projection is a dimensionality reduction technique that uses the observation that in high dimensions random vectors approach orthogonality~\cite{Ritter1989,Kaski1998}. When the columns of $\bm{R}$ are normalised to be unit vectors, $\bm{C\cdot R}$ becomes an approximately linear projection of $\bm{C}$. Another way to closely approximate normalisation of the columns of a random matrix, such as $\bm{R}$, is to sample the values of that matrix from a gaussian distribution of mean 0 and variance $\frac{1}{N}$ ($\sim\mathcal{N}\left(0,\frac{1}{N}\right)$) where $N$ is the size of the projection. This is mathematically justified by the Johnson-Lindenstrauss lemma, which states that for a set of $N$ dimensional data points there exists a linear mapping that will embed these points into an $n$ dimensional data space while preserving distances between data points within some error value, $\epsilon$. This value of $\epsilon$ is shown to decrease as $n$ increases ~\cite{johnsonLindenstrauss}
\par $RANDOM\_200$ samples from $\sim\mathcal{N}\left(0,1\right)$ also included aggregation functions (namely sum, range, and variance)~\cite{Murdock}, as discussed in \cref{implementationDetails}. It is unclear what impact this will have however preliminary investigations show little difference in performance between sampling from $\sim\mathcal{N}\left(0,1\right)$ and $\sim\mathcal{N}\left(0,\frac{1}{N}\right)$.
\par We investigate the use of random projection as an alternative to more widely used techniques by comparing each technique investigated to a random projection of the same size (\cref{fig:CaseStudies}).This should allow us to note improvements made by the quality of features as opposed to the quantity. We include \textit{RANDOM\_200} in this investigation, noting the key difference between this and the random projection being that the random numbers are drawn from different distributions (as outlined above) and that \textit{RANDOM\_200} includes aggregation functions, where a random projection does not.

\subsection{Training methods for materials science}\label{trainingMethods}
\begin{figure*}
    \centering
    \includegraphics[width=0.8\linewidth]{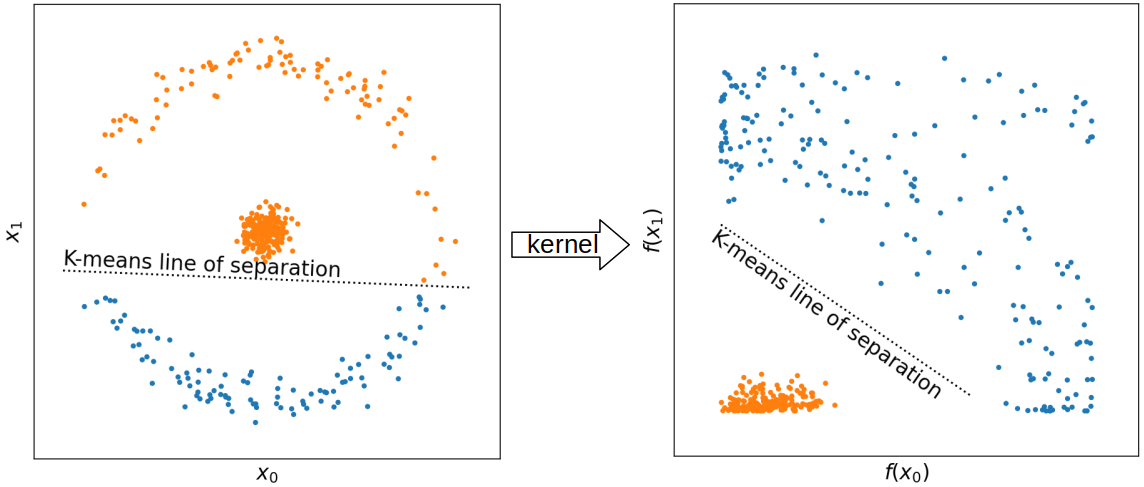}
    \caption[RBF transformation of example dataset]{A visualisation of how application of kernel functions can affect the data in an example dataset. Here we show the radial basis function (RBF) so $f(x) = \exp (-x^2) $. There is no clear way to linearly separate classes before application of RBF; however, non linear translation of each point with the RBF yields a data space through which a straight line can be drawn to separate the classes.}
    \label{fig:dummy_rbf}
\end{figure*}
Performance metrics are usually applied to a test set of data unseen by a model. Where data are scarcer, or computation time is not limiting, N-fold cross validation can be used. This is often referred to as K-fold cross validation but we use \textit{N} to avoid confusion with K-means clustering, a more central algorithm to this work. N-fold cross validation randomly splits data into \textit{N} equal sized random ``folds'', \textit{N} models are then trained, each model trained on all but one of the folds of data, and evaluated on the fold which is held out. Performance is then averaged. A common criticism of supervised ML in materials science is that datasets being worked with are inherently biased. Bias in data is a problem more broadly in ML research. In this field, exploration of similar, promising chemistries for particular applications leads to areas of the chemical data space being more dense with successfully synthesised (or DFT calculated) materials than others.
\par This leads to inflated performance metrics as performance can only be measured against other compounds that have already been synthesised (or compounds with relevant DFT calculations), as many such compounds in the test set will have similar chemistry to the training set. This can lead to comparatively poor results when trying to extrapolate to predict properties for chemistries dissimilar to those that the algorithm has been trained on. For example, it could be argued that the entries in ICSD reflect a bias towards the development of both analogues of the chemistry of minerals and chemistries lending themselves to specific types of application performance, rather than an isotropic exploration of chemical space constrained only by the inorganic chemistry of the elements themselves. Such considerations emphasise the importance of discovery synthesis that accesses new regions of chemical space, as the resulting materials can contribute to more robust models. Having robust methods to measure model performance is pertinent for materials discovery to assess likely model effectiveness in extrapolating to unseen areas of the input domain.

\subsubsection{Leave one cluster out cross validation (LOCO-CV)}\label{lococv}

\par A method to measure the extrapolatory power of an algorithm was proposed in leave one cluster out cross validation (LOCO-CV). LOCO-CV alters N-fold cross validation to have each fold contain materials in the same cluster rather than randomly selected (equal sized) folds, in order to emulate performance on unseen classes of materials.
\par Clusterings are selected using the K-means clustering algorithm~\cite{kmeans,kmeansReview}, which infers $K$ clusters ithout the need for target labels. This is done by grouping data into clusters based on their Euclidean distance to K randomly chosen “centroids". The centroids are then redefined as the mean of all points in a cluster and the data are regrouped based on these new centroids. This process is repeated until the positions of centroids (or the contents of their associated clusters) converge. K-means is quick, robust and readily implemented~\cite{skl}
\par LOCO-CV does however leave representation as a hyperparameter to the clustering (\textit{i.e.}, changing the representation will change the clusterings found with K-means clustering), and that the stochastic nature of the K-means algorithm can make measurements hard to reproduce without publishing the clusters found. A further consideration in use of LOCO-CV is that K-means does not guarantee the size of any clusters, nor does it guarantee that clusters would be deemed chemically sensible (this is discussed further in~\cref{kmeansmetrics}). It has been observed that clusters taken on materials data can vary in size by multiple orders of magnitude, which hinders the application of LOCO-CV~\cite{JonsPaper}.
\par While different sizes of clusters are to be expected in this domain (for example due to research bias in the generation of example materials), should the sizes of the clusters found in LOCO-CV differ by orders of magnitude then LOCO-CV’s ability to measure extrapolatory power is hampered. Intuitively if one of ten clusters contains 90\% of the materials in the dataset, then a measurement made with this cluster left out may give a measurement of algorithmic performance given a small fraction of the available training data, rather than indicating extrapolatory power. K-means clustering by its nature can only linearly separate clusters in a given data space. Clusters that are more distinct from one another are more likely to be isolated than clusters of data points that overlap with each other. There are other clustering algorithms, such as t-distributed stochastic neighbour embedding~\cite{TSNE}, agglomerative clustering~\cite{hierarchicalClustering}, or DBSCAN~\cite{DBSCAN}, that could be explored for LOCO-CV applications on materials datasets. We measure the separability of clusters of compounds in materials science datasets with K-means clustering.

\subsection{Kernel Methods}\label{kernelMethods}

\par While uneven cluster sizes do pose problems for LOCO-CV assessment of the extrapolatory power of ML models, such issues with K-means clustering are not solely found in materials science. K-means clustering attempts to linearly separate clusters (\textit{i.e.} draw a straight line between them), some clusters cannot be separated this way (\cref{fig:dummy_rbf}). In many cases, applying a non-linear function to every point in the dataset transforms the data in such a way that clusters can be linearly separated~\cite{primerOnKernels}. Functions used to preprocess data in this way are called kernel methods (also kernel approximation methods or kernel tricks). Prominent examples of this include RBF\cite{primerOnKernels}, additive $\chi^2$~\cite{ChiSquared}, and skewed $\chi^2$~\cite{ChiSquared}. We look at the first of these in more detail to illustrate how such kernel methods affect data points. The RBF can be defined as:
$$f(x) = \exp (-\gamma x^2)    $$

Where $\gamma$ is a hyperparameter which was set as 1 throughout this study (1 is the default for this hyperparameter in the library used). Here $x \in D $ where $D$ is a dataset of materials each represented by a feature set $R^n$ where $n$ is the dimensionality of the feature set. Examination of this formula lends intuition to the effects seen in its application (\cref{fig:dummy_rbf}), but also highlights that this function does distort the geometry of an input data space. Thus, some analysis of the results of this function are inappropriate, such as inferring meaning from changes in distances between specific points. Despite these potential caveats, non-linear transformations (e.g., through application of kernels) are frequently used with linear discrimination (such as K-means clustering)\cite{primerOnKernels}.In this paper we investigate the effect of kernel methods such as RBF on materials science data, specifically studying the use of such methods to improve suitability of LOCO-CV by addressing the problem of uneven cluster sizes outlined in \cref{lococv}. We find RBFs reduce the variance of class sizes in a clustering, regardless of input featurisation and note that this results in more reliable model convergence when using these clusterings for LOCO-CV.
\begin{figure*}
    \centering
    \includegraphics[width=0.5\linewidth]{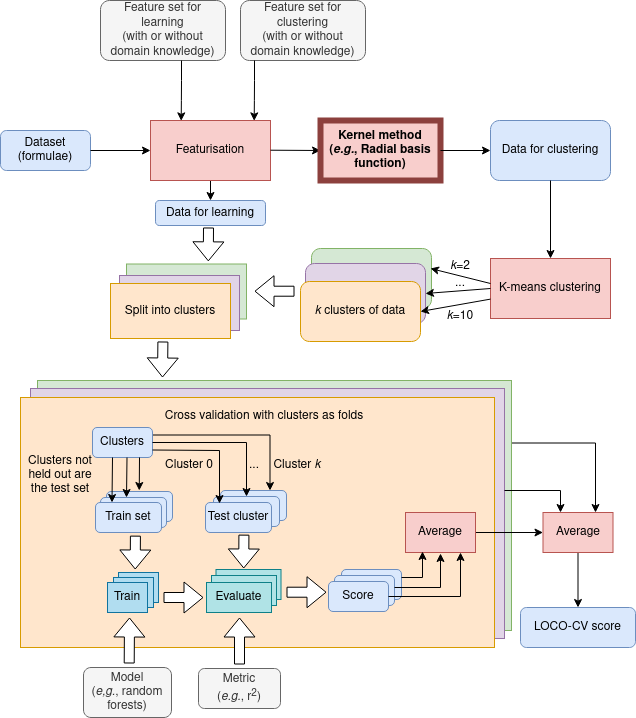}
    \caption{A flow chart demonstrating the kernelised LOCO-CV process}
    \label{fig:flowChart}
\end{figure*}
\subsection{Performance metrics in K-means clustering}\label{kmeansmetrics}
\par Without prior knowledge of expected clusters for each data point, results found with K-means clustering are difficult to interpret, though expert inspection can yield insights into what different clusters can represent. Expert inspection of results may be justifiable with less than 10 clusters (each of which could have thousands of materials), however, when using K between 2 and 10 (as was originally proposed~\cite{LOCOCV}), the LOCO-CV algorithm presents 54 different clusters ($\sum^{10}_{n=2}n $), making such expert inspection infeasible. Thus metrics must be used to quantify the success of a clustering.
\par Where target labels exist, metrics such as mutual information score, homogeneity, and completeness scores can be used. Without labels, Euclidean distance-based measures such as sum-squared distance to cluster centroid or average distance between each point and the other points in its cluster can be used, however this does not intrinsically tell us how much information is in a clustering, just how tightly packed a cluster’s members are. The average distance between each point and the other points in its cluster is computationally prohibitive so will not be used in this study.
\par Euclidean distance-based measurements such as these lack comparability in our use case, as each dataset and each featurisation technique should be considered independent. Identifying trends in these measurements with different numbers of clusters and looking at the effect of kernel methods on Euclidean distance-based measurements are both valid uses. However, as Euclidean space is affected by dimensionality, it is important that conclusions into the effect of different featurisation approaches are not drawn from such measures. While noting these caveats, we use the mean distance of a point in a cluster to the cluster’s centroid as a measure of how tight the clusters are in Euclidean space, we label this metric the spread of cluster.
\par As the aim of this investigation is to improve the validity of measures taken with LOCO-CV, specifically to address issues with vastly uneven cluster sizes, we also use the standard deviation in cluster sizes as a metric for success (the unevenness in cluster sizes). Material science datasets may have uneven cluster sizes due to research bias towards exploration of promising materials, and identically sized clusters would be unexpected for materials data, identically sized clusters were, in practice, never observed in this study. Using the unevenness of cluster sizes serves as a measure of whether cluster sizes differ by many orders of magnitude, which would affect the validity of measurements taken using LOCO-CV. This does not imply that more even clusters are more chemically sensible groupings of materials, just that they may be more sensible for use with LOCO-CV, as uneven cluster sizes bring into question measurements taken with LOCO-CV (\cref{lococv}). 
\par Unlike spread of cluster, it is valid to compare standard deviation in cluster sizes between featurisation techniques, however different datasets would be expected to differ in their ease of clustering. As such, we perform max-min normalisation across different featurisation techniques and numbers of clusters in the same dataset, and use these normalised figures to compare between datasets.

\section{Results}

\begin{figure}

\centering
\begin{subfigure}{0.65\linewidth}
  \centering
  \includegraphics[width=\linewidth]{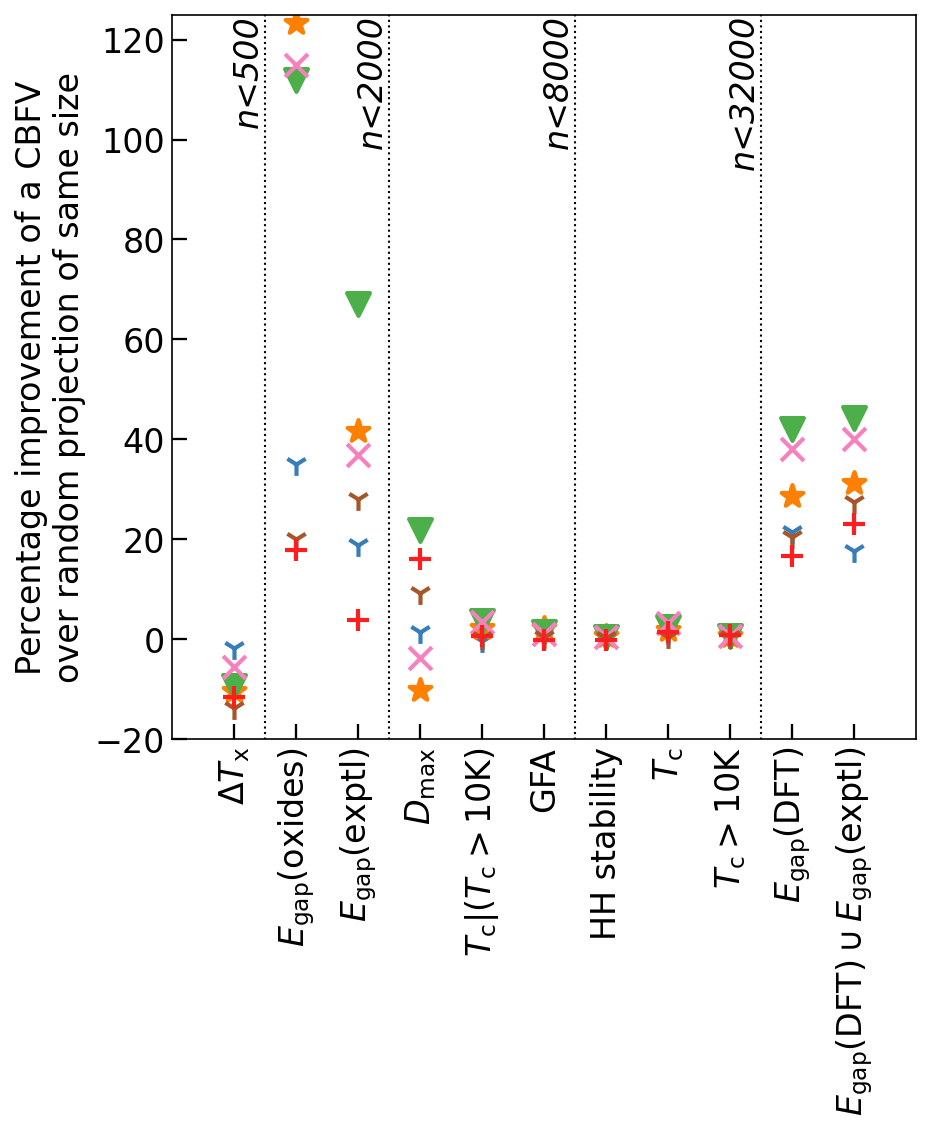} 

  \label{fig:CBFVPerformanceEqualSize}
\end{subfigure}
\begin{subfigure}{0.34\linewidth}
\raisebox{20mm}{\includegraphics[width=\linewidth]{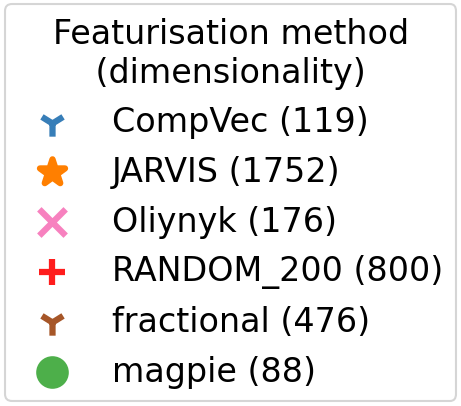}}
\end{subfigure}
\caption[Performance of composition based feature vectors on predictive tasks compared to random projections]{Performance of composition-based feature vectors (CBFVs) on predictive tasks compared to random projections.Random projections exhibit similar performance to CBFVs for most tasks. This is not true for band gap prediction tasks, where CBFVs with domain knowledge demonstrate marked improvement.}
\label{fig:CaseStudies}
\end{figure}
\subsection{Effect of representation on predictive ability of random forest: Case Studies}~\label{caseStudies}
We examine five case study publications’ datasets to compare the representations used in them with a non-structural CBFV examined in previous work~\cite{Murdock}, and with the composition vector (CompVec) suggested for use with ElemNet~\cite{elemnet}. Case studies have been selected to incorporate the prediction of a variety of material properties, research groups, and notable works that reflect the state-of-the-art. We use the original datasets to replicate studies, but use 80-20 train-test splits. 
\par We use a consistent 80-20 train-test split across all data sets to enable us to draw conclusions about which representations work better generally. This should help us to establish whether previous findings (i.e. that domain knowledge is more beneficial in smaller datasets and that benefit diminishes as dataset size increases over 1000)~\cite{Murdock}, hold true for RFs. LOCO-CV measurements for these experiments are available in the supporting information, and the clusterings found for LOCO-CV are available in the associated git repository~\cite{myGit}. 
\par Representations compared are:
\begin{itemize}
    \item \textit{Oliynyk}~\cite{Oliynyk}. Originally designed for prediction of Heusler structured intermetallics~\cite{Oliynyk}, the \textit{Oliynyik} feature set as implemented in previous work includes 44 features~\cite{Murdock}. For each of these, the weighted mean, sum, range, and variance of that feature amongst the constituent elements of the compound are taken. Features include atomic weight, metal, metalloid or non metallic properties, periodic table based properties (Period, group, atomic number), various measures of radii (atomic, Miracle, covalent), electronegativity, valency features (such as the number of s, p, d, and f valence electrons), and thermal features (such as boiling point and specific heat capacity).
    \item \textit{JARVIS}~\cite{JARVIS}: \textit{JARVIS} combines structural descriptors with chemical descriptors to create ``classical force-field inspired descriptors'' (CFID). Structural descriptors include bond angle distributions neighbouring atomic sites, dihedral atom distributions, and radial distributions, among others. Chemical descriptors used include atomic mass, and mean charge distributions. Original work generated CFIDs for tens of thousands of DFT-calculated crystal structures \cite{JARVIS}, and subsequent work adapted CFIDs for individual elements to be used in CBFVs for arbitrary compositions without known structures (\textit{i.e.}~\cref{fig:CBFVworkflow})~\cite{Murdock}.
    \item \textit{Magpie}~\cite{Ward2016}: While the Materials-Agnostic Platform for Informatics and Exploration (MAGPIE) is the name of a library associated with Ward \textit{et al.}’s work, it this has become synonymous with the 115 features used in the paper and as such we will use \textit{Magpie} refer to the feature set. These features include 6 stoichiometric attributes which are different normalistion methods ($L^P$ norms) of the elements present. These capture information of the ratios of the elements in a material without taking into account what the elements are, 115 elemental based attributes are used, which are derived from the minimum, maximum, range, standard deviation, mode (property of the most prevalent element) and weighted average of 23 elemental properties including atomic number, Mendeleev number, atomic weight among others. Remaining features are derived from valence orbital occupation, and ionic compound attributes (which are based on differences between electronegativity between constituent elements in a compound).
    \item \textit{RANDOM\_200}~\cite{Murdock}: a random vector featurisation used by Murdock~\textit{et al.} to represent a lower bounds for performance.
    \item  \textit{fractional}\cite{Murdock}: An implementation of a one-hot style encoding of composition which includes average, sum, range, and variance of each element.
    \item \textit{CompVec} a one-hot style encoding of composition as used in ElemNet~\cite{elemnet} (containing only the proportions of each element in a composition). Differences between this and \textit{fractional} are further discussed in \cref{representations}.
\end{itemize}

We compare each of these representations to a random projection of equal size. This allows us to control for the size of a representation when investigating the advantage of the domain knowledge built into a CBFV. Several of the five case studies investigated contain multiple applications of ML within a single publication. The tasks which were recreated in this comparison (and their relevant case study references) are as follows: 
\begin{itemize}
    \item $T_\mathrm{c}$: Using a regressor to predict the superconducting critical temperature ($T_\mathrm{c}$) of a material (12666 data points in training set) ~\cite{Stanev}.
    \item  $T_\mathrm{c}>10\mathrm{K}$: Classifying if the $T_\mathrm{c}$~of a material is greater than 10K (12666 data points in training set)~\cite{Stanev}.
    \item $T_\mathrm{c} | (T_\mathrm{c}>10\mathrm{K})$: Regressing to find $T_\mathrm{c}$~given $T_\mathrm{c}>10\mathrm{K}$~K (4833 data points in training set)\cite{Stanev}.
    \item HH stability: Predicting the stability of half-Heuslers (8948 data points in training set) ~\cite{Legrain2018}.
    \item $E_\mathrm{gap}(\mathrm{oxides})$: Predicting the band gap of oxides found in the Computational Materials Repository database (599 data points in training set)~\cite{Davies2019}.
    \item Glass Forming Ability (GFA): predicting the ability of a bulk metallic glass alloy (BMG) to exist in an amorphous state (5051 data points in training set)~\cite{Ward2018}.
    \item $D_\mathrm{max}$: Predicting the critical casting diameter of a BMG (4724 data points in training set)~\cite{Ward2018}.
    \item $\Delta T_\mathrm{x}$: The supercooled liquid range of a BMG (495 data points in training set)~\cite{Ward2018}.
    \item $E_\mathrm{gap}(\mathrm{DFT})$: Predicting the band gap of materials calculated using DFT (35653 data points in training set)~\cite{Sparks2018}.
    \item $E_\mathrm{gap}(\mathrm{exptl})$: Predicting the band gap of materials measured experimentally (1986 data points in training set) ~\cite{Sparks2018}.
    \item $E_\mathrm{gap}($DFT$) \cup E_\mathrm{gap}($exptl$)$: Predicting the band gap of a dataset consisting of both DFT calculated and experimentally measured band gaps (37639 data points in training set)~\cite{Sparks2018}.
\end{itemize}
We report measured performance in regression tasks was using $r^2$ correlation and classification task performance is measured using accuracy. Thus percentage improvement over random projections can be considered to be: 
$$100\left( \frac{ M(y, \hat{y})} { M(y, \hat{y}_p)} -1\right) $$ 
Where $y$ is the target label for a prediction, $\hat{y}$ is the label predicted by a model that uses a given representation, $\hat{y}_p$ is a label predicted by a model that uses a random projection of equal size to the given representation, and $M$ is accuracy for classification tasks and $r^2$ for regression tasks. Measurements found using other values of $M$ can be found in the supplementary information.

\begin{figure*}
\centering
\begin{subfigure}{70mm}
  \includegraphics[width=70mm]{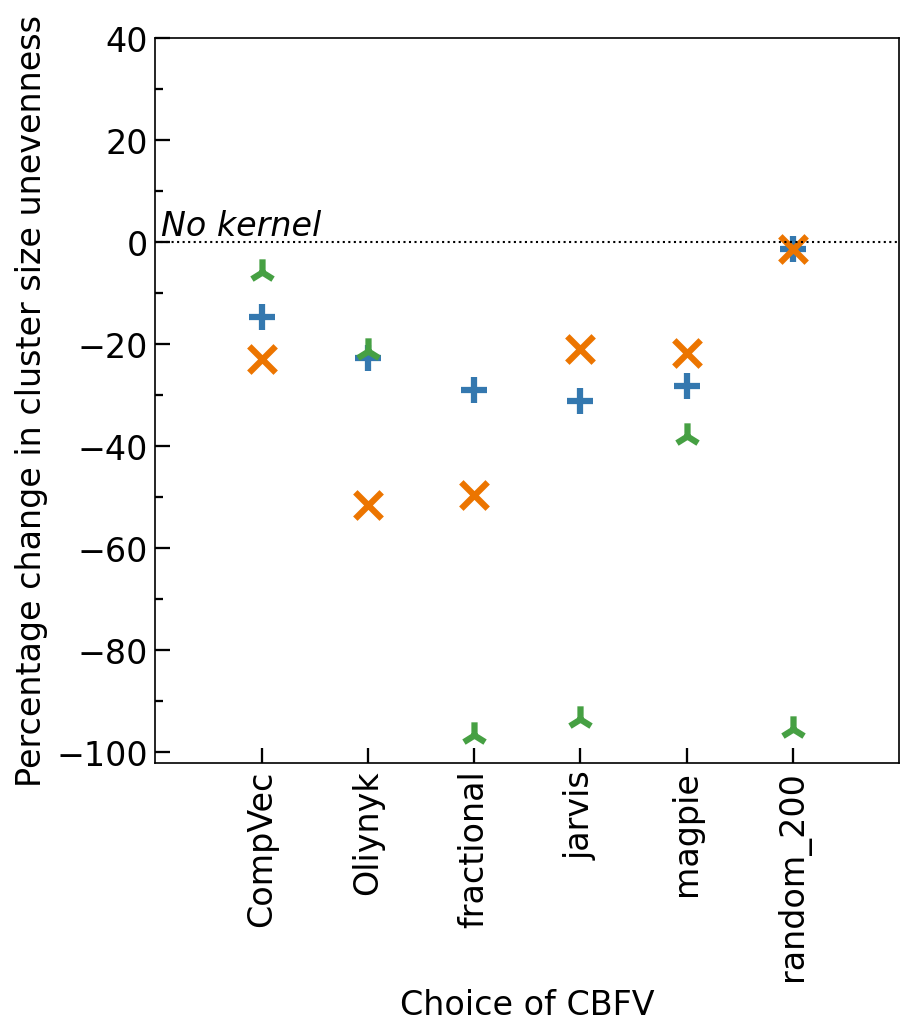}
  \caption{}
\end{subfigure}\hfill
\begin{subfigure}{94mm}
  \includegraphics[width=70mm]{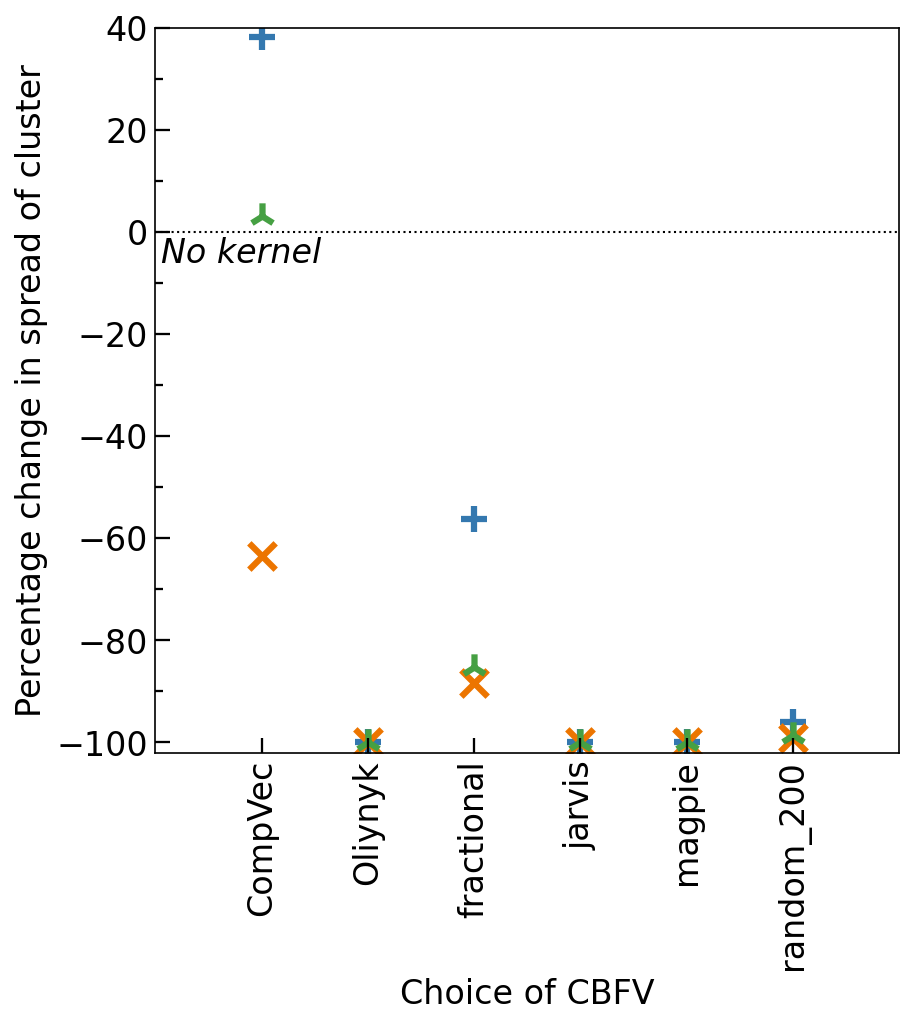}
  \raisebox{40mm}{\includegraphics[width=23mm]{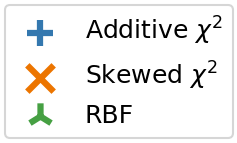}}
  \caption{}
\end{subfigure}

\begin{subfigure}{75mm}
\includegraphics[width=75mm,trim={0 1.25cm 0 1.1cm},clip]{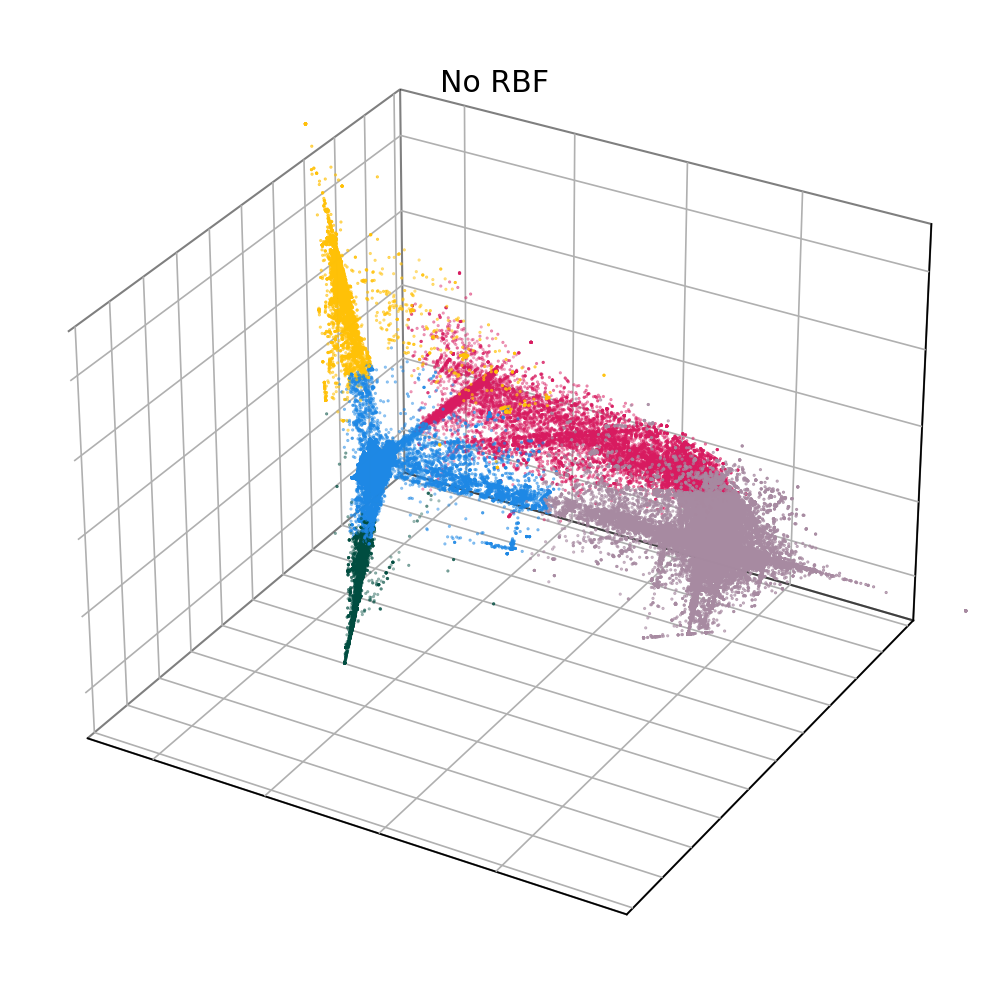}
    \caption{}
    \label{fig:PCAProjection}  
\end{subfigure}
\hfill
\begin{subfigure}{94mm}
\includegraphics[width=75mm,trim={0 1.25cm 0 1.1cm},clip]{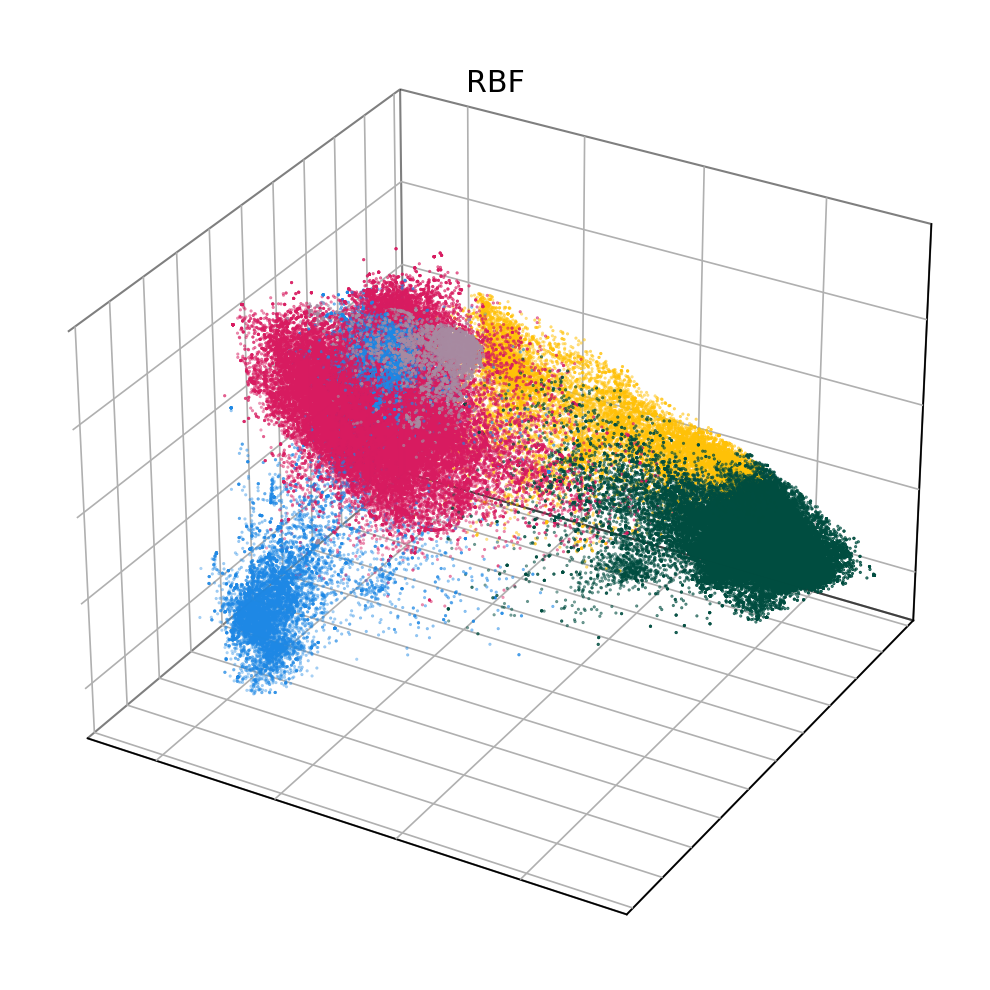}
    \caption{}
    \label{fig:PCAProjection2}  
\end{subfigure}

\caption[Demonstration of kernel methods on clustering of compositions in the ICSD.]{Demonstration of the effect of kernel methods on clustering of compositions in the ICSD. 
(a) Changes in standard deviation of cluster size found by K-means clustering of ICSD (k=5) with application of kernel methods. Most of the time, application of kernel methods reduces the variation between cluster sizes. This effect is most pronounced with the basis function (RBF) kernel. 
(b) Variation in cluster spread for K-means clustering of ICSD (k=5). Application of kernel methods reduces the spread in Euclidean space within a cluster. This effect is most pronounced with skewed $\chi^2$ and RBF. 
(c) To visualise these results, PCA was used to generate the first three principal components of all compositions in the ICSD featurised using a CompVec. Colours correspond to clusters found by K-means (k=5) clustering on this representation. Inspection of these clusters reveals highly anisotropic clusters with no meaningful boundaries in the data to unambiguously separate clusters. 
(d) The first three principal components found when examining an RBF translation of the ICSD (featurised using CompVec), points are coloured according to clusters found by K-means (k=5) applied to the kernelised data. The application of an RBF (as defined in~\cref{kernelMethods}) to every composition vector in the ICSD (before clustering) leads to clusters that are more isotropic with more clearly resolved boundaries between clusters.}
\label{fig:ICSDPerformance}
\end{figure*}
\begin{figure*}
    \centering
    \begin{subfigure}{70mm}
  \centering
  \includegraphics[width=70mm]{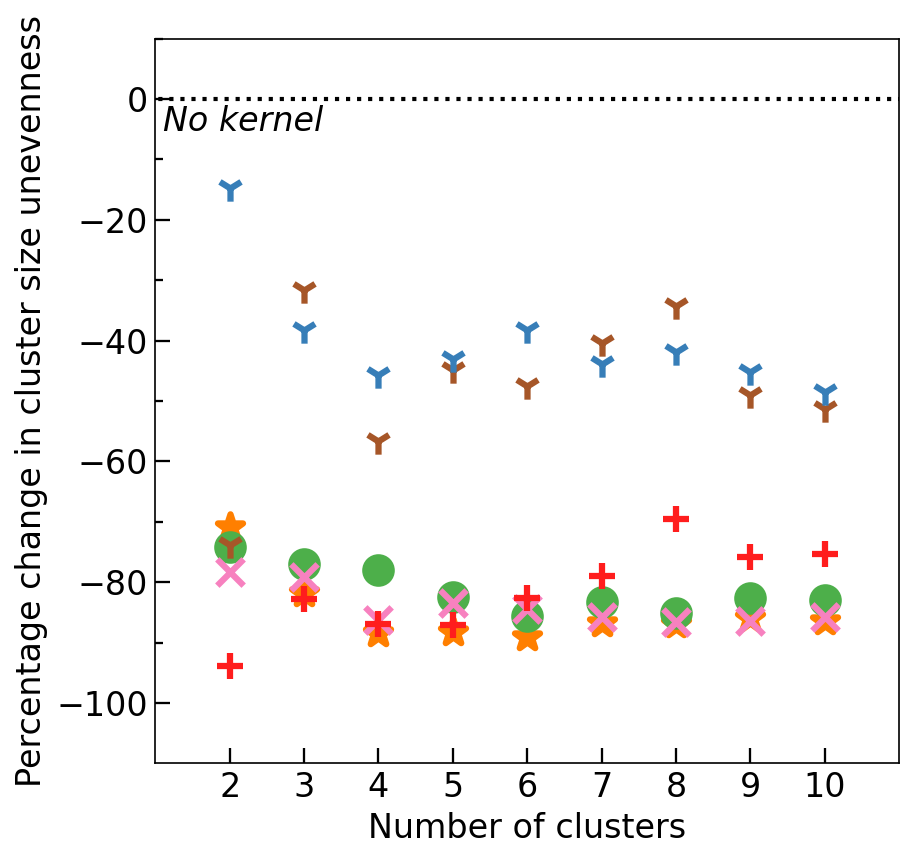}
  \caption{}
  \label{fig:stdVsRBF}
\end{subfigure}\hfill
\begin{subfigure}{70mm}
  \centering
  \includegraphics[width=70mm]{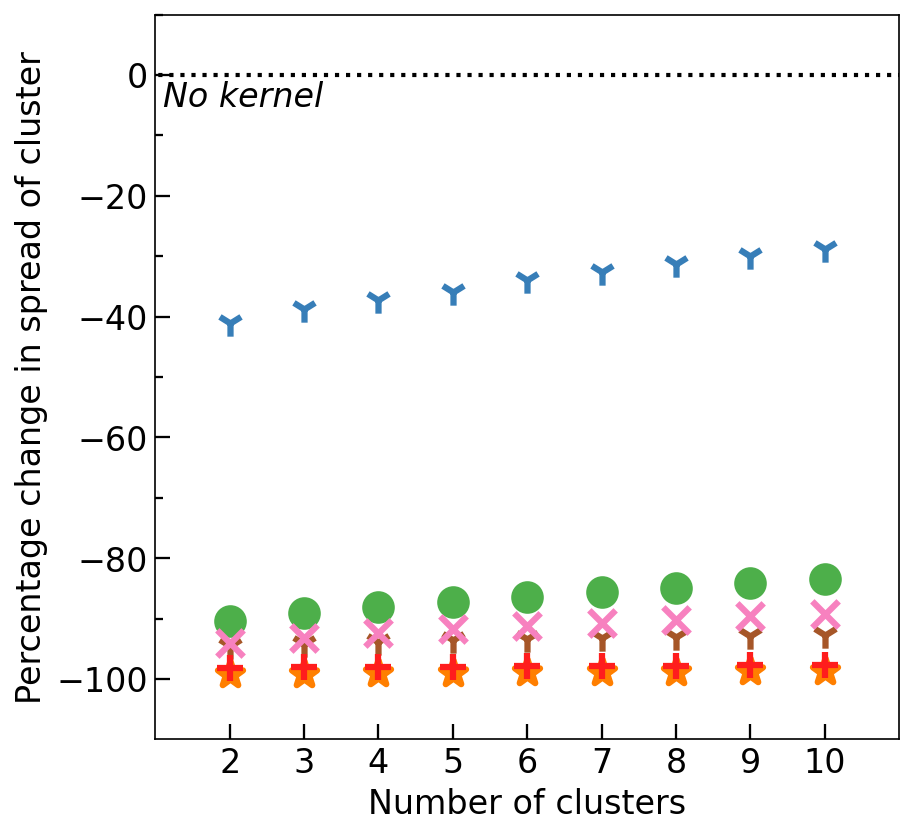}  
  \caption{}
  \label{fig:distVsRBF}
\end{subfigure}
\begin{subfigure}{30mm}
  \includegraphics[width=\linewidth]{fig9aLegend.png}
\end{subfigure}
\caption[Effect of radial basis function (RBF) on clustering]{Effect of radial basis function (RBF) on standard deviation of cluster sizes (cluster size unevenness) and spread of cluster sizes. This is performed using K-means clustering with different values of $k$. 
(a) RBF leads to more evenly sized clusters for all featurisation methods and nearly all values of $k$. 
(b) RBF leads to more compact clusters (i.e., smaller average Euclidean distance between points within a cluster) for all featurisation methods and all values of $k$}\label{fig:generalKernelPerformance}

\end{figure*}
\par Overall, recreation of these tasks shows that, broadly, changes in CBFV made little difference to performance when compared to a random projection of the same size (\cref{fig:CaseStudies}). Featurisation methods inspired by domain knowledge do show advantages in some datasets. These advantages seem to be task-specific as opposed to based on dataset size, specifically band gap-based tasks seem to see benefit from knowledge-based features, however most other tasks do not see noticeable improvement from this feature engineering (\cref{fig:CaseStudies}). This could be because vast amounts of band gap data can be acquired through DFT calculations~\cite{materialsProject}~and as such band gap prediction is a widely available benchmark that researchers could use when testing a newly proposed CBFV\cite{materialsBenchmarks}. 
\par Intuition may suggest introducing more dimensions that do not contain any additional information would result in worse algorithmic performance. However, despite having 68\% more dimensions, RANDOM\_200 performs within 5\% of the fractional representation. On large enough data sets ($\sim 3000 < n$) the random representation does not perform appreciably differently to the Magpie representation. Notably on tasks outside of band gap prediction there is little advantage to domain based representations over a random projection. 
\par We encourage the use of random projection as an alternative to CBFV, and propose its use as a comparative measure against CBFV. If a feature set cannot appreciably outperform a random projection of the same size or smaller, then, while there may still be benefits to analysis of the feature importance of such a feature set, that feature set does not enrich the representation of a material when it comes to algorithmic performance.

\subsection{Improving the linear separability of chemical data spaces for more applicable measurements of extrapolatory power}\label{linearSeperability}
\begin{figure*}
    \centering
    \begin{subfigure}{70mm}
  \centering
  \includegraphics[width=70mm]{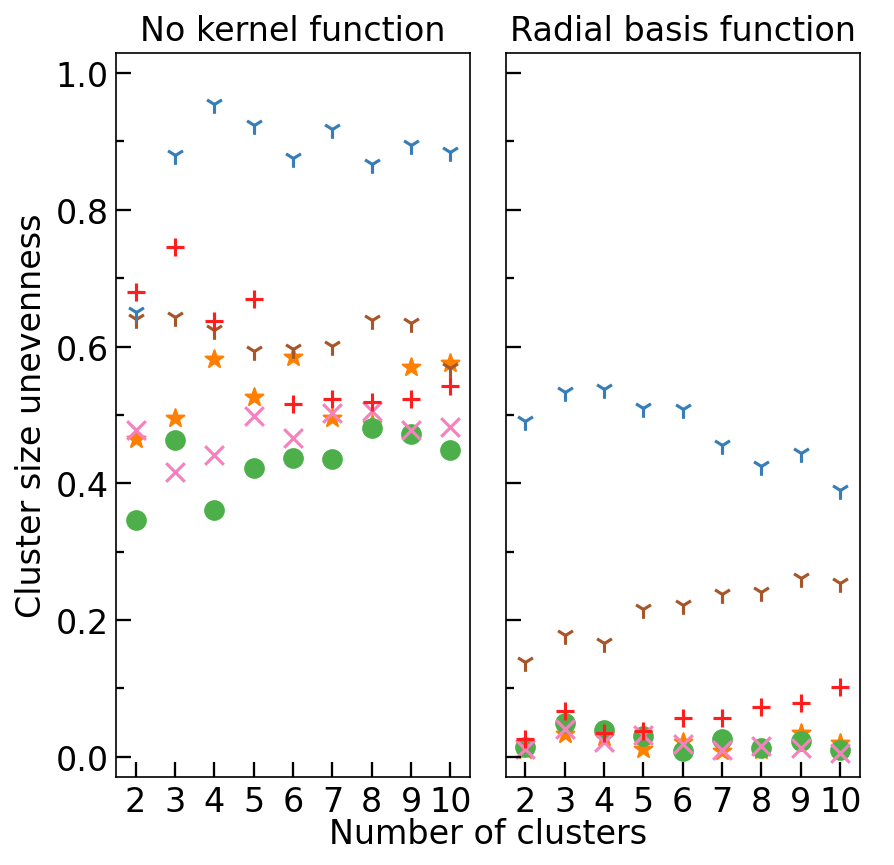}
  \caption{}
  \label{fig:stdVsRBFAbsolute}
\end{subfigure}\hfill
\begin{subfigure}{70mm}
  \centering
  \includegraphics[width=70mm]{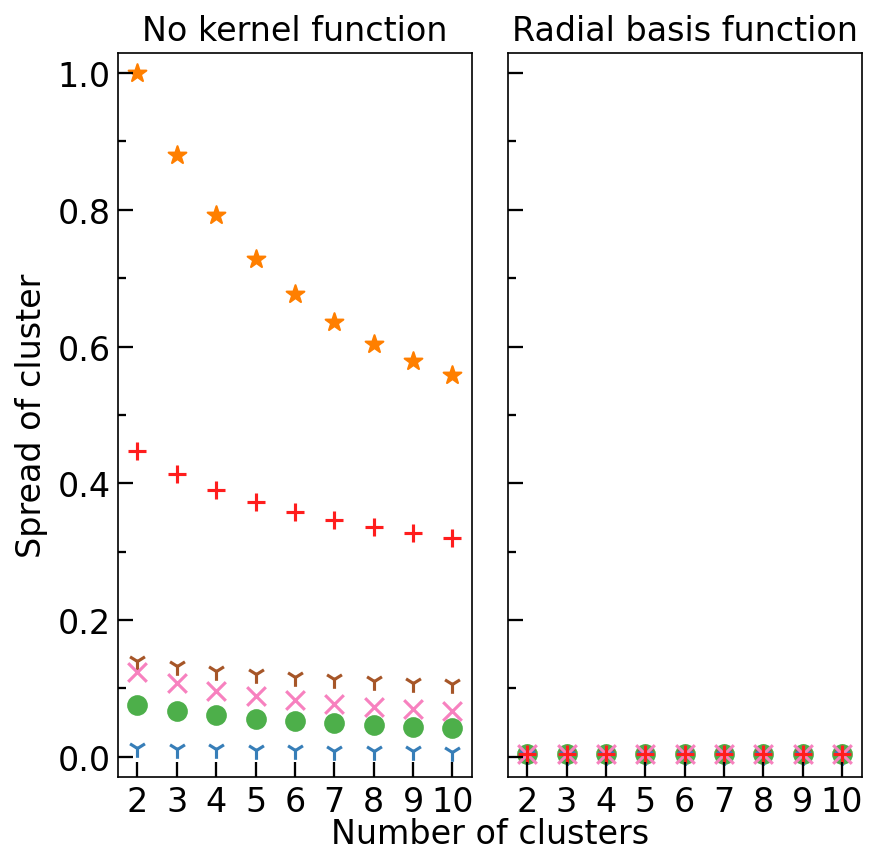}  
  \caption{}
  \label{fig:distVsRBFAbsolute}
\end{subfigure}
\begin{subfigure}{30mm}
\includegraphics[width=\linewidth]{fig9aLegend.png}
\end{subfigure}
\caption[Average cluster size evenness and spread of clusters when clustering different representations of datasets]{Mean cluster size unevenness and spread of clusters found by K-means when clustering different representations of datasets. Measurements are normalised to between one and zero on a per dataset basis, as different datasets would be expected to cluster with different amounts of ease. The normalised values are then averaged across different datasets for each representation and value of $k$. 
(a) Clusters are generally more even in domain knowledge based representations as measured by the standard deviation in cluster sizes. 
(b)Without application of kernel function, spread of clusters as measured by the average distance between a point in a cluster and its centroid correlates to the size of the representation with the exception of \textit{CompVec} which has the tightest clusters. Application of radial basis function makes this trend insignificant.}
\end{figure*}

\par We investigated which of the representations of a compound outlined in~\cref{caseStudies}~will lead K-means clustering to identify more evenly sized clusters in different datasets. Datasets investigated were those used in~\cref{caseStudies}~as well as the inorganic crystal structures database (ICSD) as a whole.
\par In classical computer science problems, non-linear kernels have been applied to datasets on which a linear discriminator (such as K-means, or support vector machines) exhibits poor performance. As described in~\ref{kernelMethods}, applying a non-linear transformation (e.g., a kernel function) to every data point in a data set can transform data such that it is more amenable to linear discrimination ( (\cref{fig:dummy_rbf}). We applied the radial basis, additive $\Tilde{\chi}^2$, and skewed $\Tilde{\chi}^2$ functions to the investigated representations to see if these non-linear translations will reduce cluster size unevenness found by K-means clustering. Reduced cluster size unevenness found with K-means would improve the applicability of LOCO-CV measurements, addressing one of the problems highlighted in~\cref{trainingMethods}.
\par As additive $\Tilde{\chi}^2$, and skewed $\Tilde{\chi}^2$ functions are only well defined for positive inputs, data was scaled between 0 and 1 using min-max normalisation before these methods were applied. As RBF (and K-means without kernels) can be affected by disparity of scale between axis to check to investigate the effects of different normalisation methods were investigated, with the data normalisation which most often resulted in the lowest cluster size uneveness being used for the results below (no normalisation was used with RBF and min-max scaling to between -1 and 1 was used when no kernel method was being applied). Further details of this can be seen in section S1 of the supplementary material.
 \par All three kernel functions investigated resulted in more evenly sized clusters than no kernel function being applied at all, with RBF, on average, resulting in the largest reduction in standard deviation between cluster size (\cref{fig:ICSDPerformance}). Additionally, we note that application of any of these kernel methods generally resulted in a reduction in distance between points in a cluster and their centroids (spread of cluster), indicating more tightly packed clusters (\cref{fig:distVsRBF}). On average application of skewed $\Tilde{\chi}^2$ saw the greatest reduction in spread of cluster. As this investigation looks to create more even cluster sizes for use with LOCO-CV we focus on impacts of RBF, as, of the kernel methods tested, it resulted in the greatest impact on this metric as defined by the largest reduction in standard deviation of cluster size. 
\par Before application of a kernel function, we note that cluster sizes are more even in domain knowledge-based representations as measured by the standard deviation in cluster sizes. \textit{CompVec} representation resulted in a larger standard deviation between cluster sizes (i.e., less evenly sized clusters) than all other representations investigated, likely due to the sparse nature of this representation, with the \textit{magpie} representation resulting in the most even cluster sizes (\cref{fig:stdVsRBFAbsolute}). The two one-hot based representations, \textit{fractional} and \textit{CompVec}, generally did not result in as even cluster sizes as other representations. Application of \textit{CompVec} resulted in performance substantially worse than that of \textit{fractional} despite them being very similar nature, only differing in use of aggregation functions (as discussed in~\cref{representations}).
\par RBF universally resulted in more even clusters. The smallest change (as a percentage of the standard deviation in cluster size before application of RBF), was seen in \textit{fractional} and \textit{CompVec} representations (two of the representations which resulted in the worst performance in this metric) (\cref{fig:stdVsRBF}). However, outside these two representations, the proportional impact of RBF on this measure did not correlate to the performance of a CBFV in this measure prior to application of RBF.

\begin{figure*}
\centering
\begin{subfigure}{83mm}
\includegraphics[width=76mm]{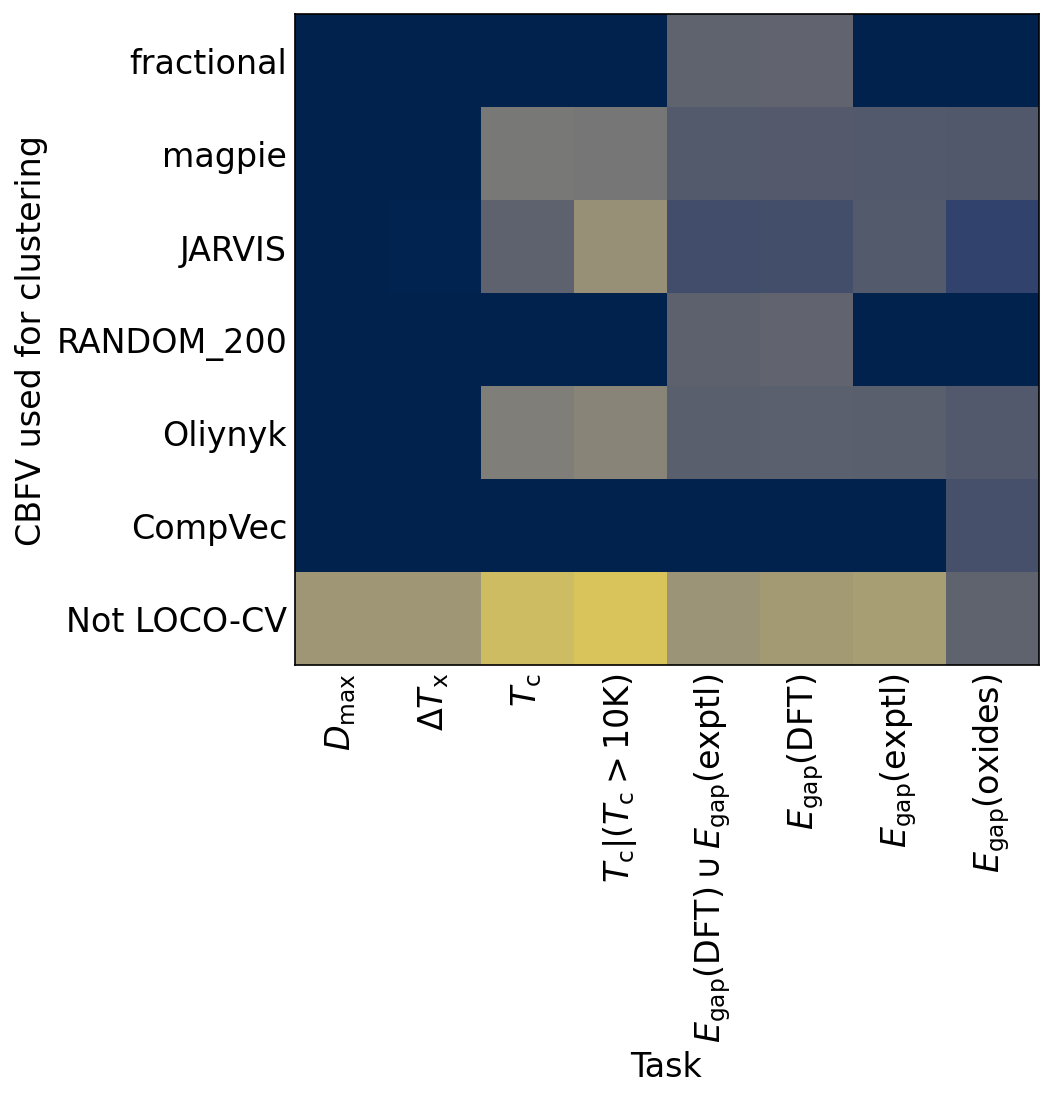}

\caption{}
\label{fig:LOCOCVr2AcrossRepresentations}
\end{subfigure}
\begin{subfigure}{90mm}
\includegraphics[width=90mm]{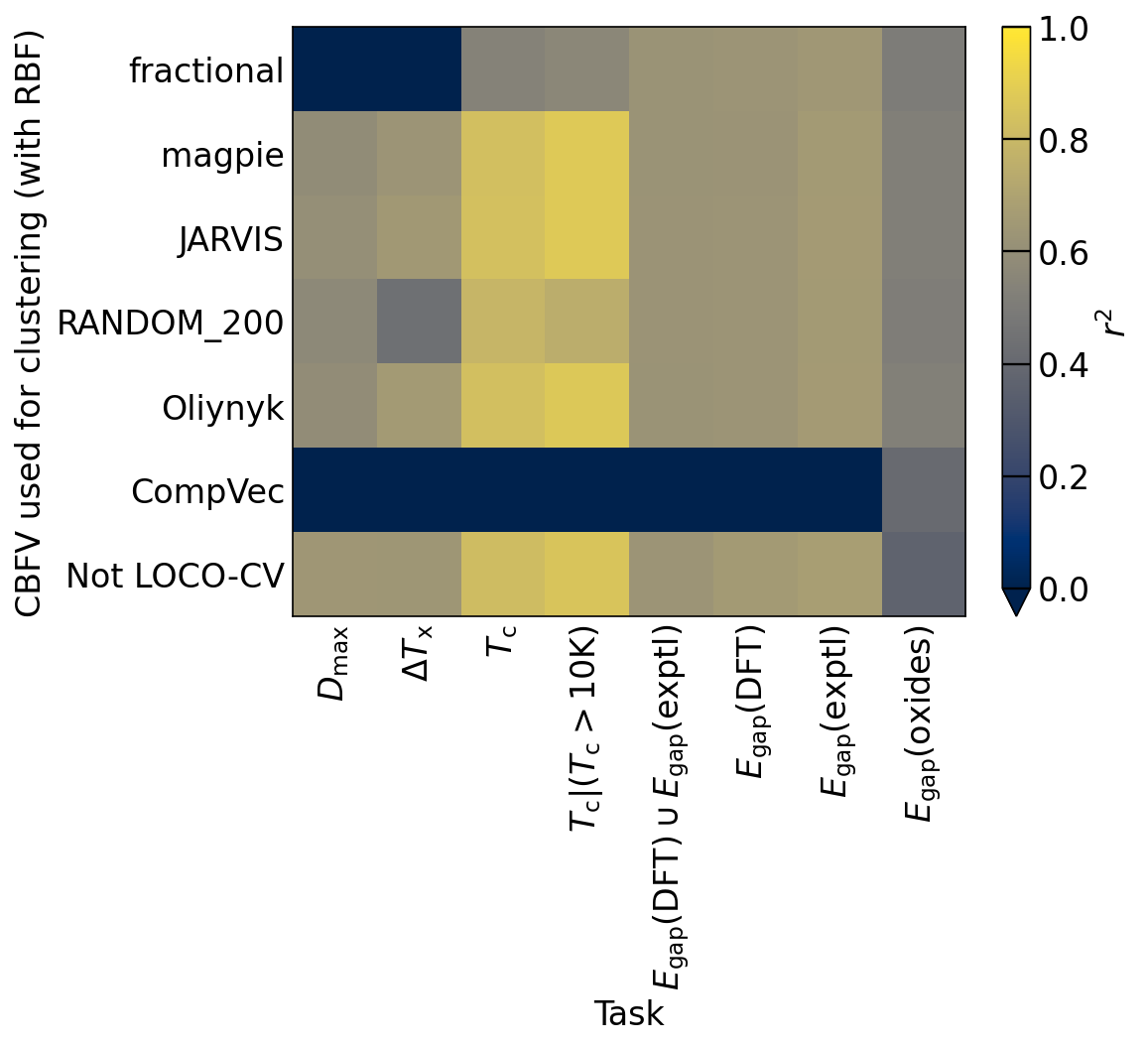}
\caption{}
\label{fig:KernelisedLOCOCVr2AcrossRepresentations}
\end{subfigure}

    \caption{(a) $r^2$ performance of regression tasks measured in LOCO-CV clustered using different CBFVs, and measured using traditional 80-20 split (labelled as ``Not LOCO-CV"). In all cases the model was trained using data in a CompVec representation, so we can better examine the effect of LOCO-CV on the measurement drawn from a given model. We see in many of the cases the same model which performs well in traditional 80-20 split training regimen fails to converge in LOCO-CV measurements. (b) Application of RBF to CBFVs before K-means clustering for LOCO-CV results in much fewer models failing to converge than those seen in (a)}
    \label{fig:LOCOCVConvergence}
\end{figure*}

\par Without use of kernel functions, there is a clear correlation between the size of a representation and the spread of the clusters found using that representation, with the exception of \textit{CompVec}, which saw the tightest clusters  (\cref{fig:distVsRBFAbsolute}). This trend is no longer seen after application of RBF. Application of RBF to a CBFV before K-means clustering reduced the spread of clusters found (\cref{fig:distVsRBF,fig:distVsRBFAbsolute}). The relative size of the change seen after application of RBF correlated with the spread of clusters found when no kernel method was used. The higher the spread of clusters found using a CBFV without a kernel method, the larger the change seen when clustering using that CBFV and a RBF.
\par Use of kernel methods in featurisation results in more even cluster sizes when using that featurisation for K-means clustering. As featurisation used for clustering in LOCO-CV is independent of that used for learning, incorporating these kernel methods into LOCO-CV is simple and applicable regardless of machine learning algorithm, chosen metric, and initial representation (\cref{fig:flowChart}). Thus we recommend use of kernel methods when using K-means clustering for LOCO-CV to address the issue of uneven cluster sizes (as discussed in \cref{trainingMethods}). Addressing this issue results in models being more successful at converging using LOCO-CV (\cref{fig:LOCOCVConvergence}), and adds applicability to measurements taken with LOCO-CV, allowing for better measurements of the extrapolatory power of an algorithm, which is of particular importance in materials science. 
\subsection{Clustering Random Projections with and without kernel methods}\label{clusteringRandomProejctions}
Having established that random projections perform similarly to engineered feature vectors in many task (\cref{caseStudies}) and that kernel methods can be used to reduce cluster size variance in K-means clustering on materials datasets (\cref{linearSeperability}), experiments were carried out to measure the cluster size variance of random projections of compositions both with and without application of kernel methods. 
\par
\begin{figure*}
\centering
\begin{subfigure}{92mm}

\includegraphics[width=65mm]{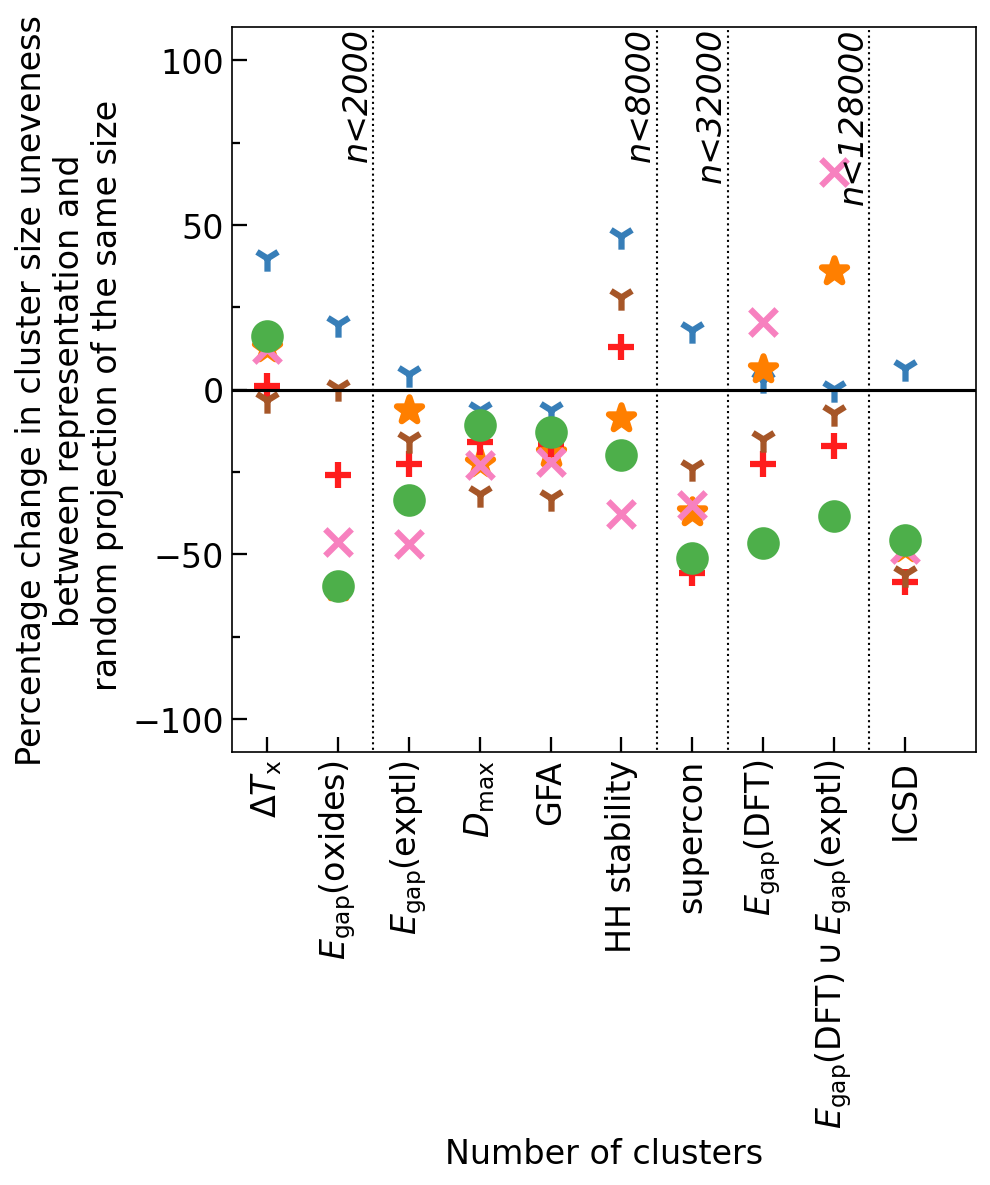}
\raisebox{40mm}{\includegraphics[width=26mm]{fig9aLegend.png}}

\caption{}
\label{fig:projectionVsRepresentation}
\end{subfigure}
\begin{subfigure}{83mm}

\raisebox{21mm}{\includegraphics[width=64mm]{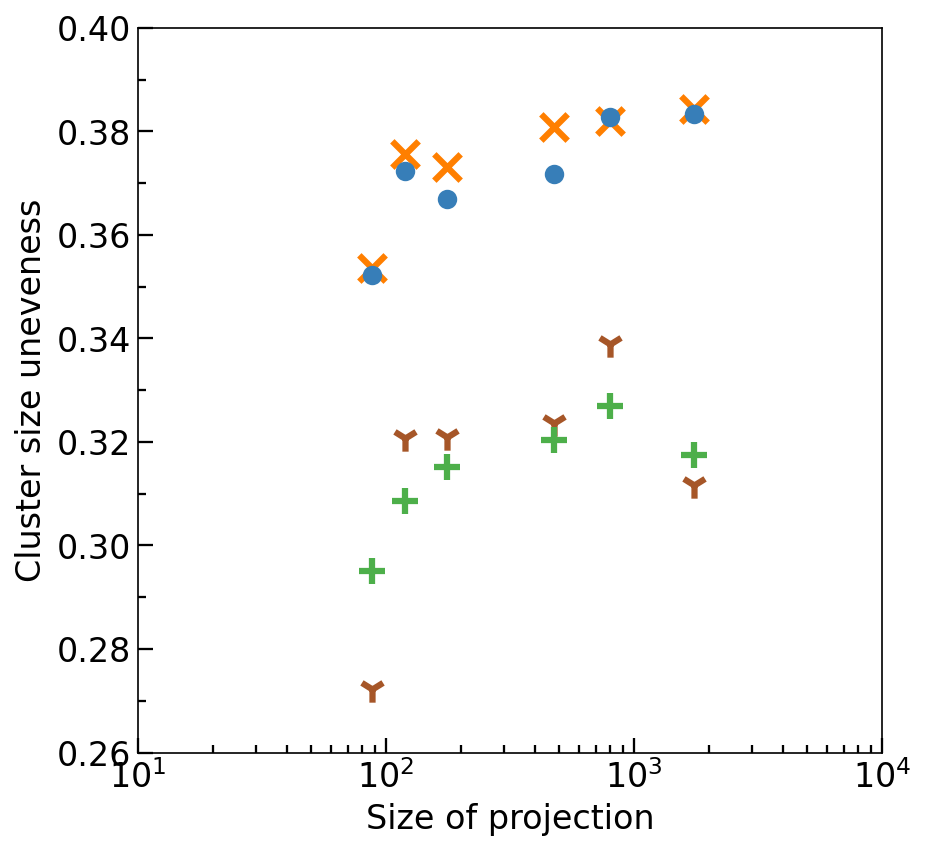}}
\raisebox{40mm}{\includegraphics[width=18mm]{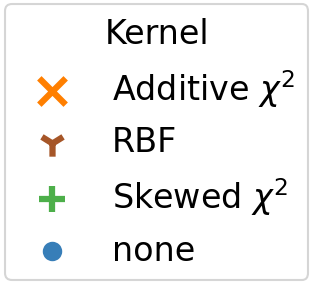}}

\caption{}
\label{fig:projectionSizeVsUnevenes}
\end{subfigure}

    \caption{(a) Reduction in cluster size unevenness (standard deviation in cluster size) of different CBFVs when compared to equal sized random projections of composition vectors across different datasets with no kernel applied. While random projection consistently outperformed \textit{CompVec}, all other CBFVs form more even clusters than an equally sized random projection. (b) Average cluster size unevenness found using K-means clustering on datasets featurised using random projections of various sizes. Cluster size variances are normalised between 1 and 0 for each dataset (as different datasets would be expected to cluster with different amounts of ease), and then averaged for each size of random projection and each kernel. RBF and skewed $\chi^2$ is seen to reduce cluster size uneveness, with the projections of approximately 100 dimensions performing better than larger projections. }
    \label{fig:RandomProjectionUneveness}
\end{figure*}

 Without application of kernel functions, when each CBFV was compared to a random projection of equal size (\cref{fig:projectionVsRepresentation}), using random projections of composition vectors did, more often than not, result in more evenly sized clusters than \textit{CompVec}, but less evenly sized clusters than all other CBFVs investigated. However, no representation (either random projection or CBFV) universally resulted in more even clusters. Comparing the best performing size of random projections (88 dimensions) with other CBFVs without any kernel methods did narrow the differences in cluster size uneveness (fig S2b), however other CBFVs still outperformed random projections in several datasets. 
\par Radial basis, additive $\chi^2$, and skewed $\chi^2$ functions were applied to these projections before clustering using K-means. The resulting clusters were compared to those found without any kernel methods, showing that RBF and skewed $\chi^2$ did reduce cluster size unevenness (~\cref{fig:projectionSizeVsUnevenes}). However, these results still do not create a consistent pattern of either outperforming or underperforming the cluster size unevenness found by applying RBF to CBFVs (fig S2a). As no representation universally results in more even clusters, a variety of CBFVs and random projections should be investigated when choosing the best representation for clustering a dataset. Application of kernel methods such as RBF are advantageous in this context regardless of representation.

\section{Discussion}
Recreation of studies discussed in \cref{caseStudies} shows that, broadly speaking, featurisation methods used in research are not necessarily advantageous over random projections, especially on tasks that are not related to band gaps. Machine learning led research in materials science often aims to highlight the success of a machine learning model either in a materials discovery pipeline, as a proof of concept that a model can learn from a given dataset, or a proof of concept that a property can be predicted. As such, the exact implementation of a CBFV and its effectiveness when compared to other CBFVs are often not included in the main text of a paper. Comparison studies thus facilitate evaluation of the impact of the CBFVs on ML performance.  
\par With modern libraries such as matminer~\cite{matminer}, creating new featurisation methods and changing existing ones is straightforward. The engineered featurisation methods show no advantage over more widely used, or simpler alternatives, in the tasks considered here. 
\par Both findings here and in previous work suggest that for sufficiently large and balanced datasets, domain knowledge in CBFVs yields only small advantage\cite{Murdock}. Promising results in representation learning could further reduce these advantages~\cite{AGoodall}, which means the question as to whether these small advantages of feature engineered CBFVs justify the difficulty in comparison between the models using them is an open one.
\par Choice of representation for a supervised ML algorithm may be influenced by the extent to which the goal of the algorithm is to maximise predictive accuracy for a property (\textit{e.g.}, to screen potential candidates for synthesis), and the extent to which the goal is to gain insight into the causes of that property. Linked to this consideration is the question of whether domain knowledge features are being used as proxy for the composition, or whether the composition is a proxy for the properties of a material which are quantified by the domain knowledge features. 
\par For example, a model trained to predict whether a superconductor has a $T_\mathrm{c}$ greater than 30K could be trained on a CBFV and find that the number of d electrons is an important indicator for this property. A similar model could be trained using a \textit{CompVec} representation and find that containing Cu is an important indicator for this property. Whether the number of d electrons is serving as proxy for the presence of Cu in a material or the presence of Cu in a material is a serving as proxy for the number of d electrons is a matter of perspective. Bearing this difference in perspective in mind may help guide towards use of a representation which is best suited for the workflow in which a machine learning algorithm is being used. If we use ML to gain insight into the causes of properties and phenomena, then examining the importance of different domain knowledge areas in a CBFV for an algorithm will allow us to do that. This would suggest that the task becomes a matter of finding the best set of properties for an element to adequately explain how it interacts with the chemistries of a compound. At this point experimenting with various combinations of elemental properties becomes appealing. However, to justify this approach adequate analysis of which properties are important is needed.
\par When choosing a representation to maximise predictive accuracy, domain knowledge seems to provide some advantage for some tasks examined here (particularly band gap prediction tasks). However we do not think this evidence, nor that found in previous work~\cite{Murdock}, is sufficient to reject featurisation methods without domain knowledge such as fractional encoding of composition or random projections, for more complex or parameter dependant algorithms. When using a CBFV, random projection offers a helpful baseline for performance as it is simple to implement and works fairly well. Their single hyperparameter is the size of the projection, which allows one to draw conclusions as to the usefulness of a CBFV under investigation without introducing the size of a representation as a contributing factor for its performance.
\par Extrapolatory power is particularly pertinent in the materials discovery field, thus previous work presented LOCO-CV as a way to estimate the extrapolatory power of a supervised machine learning algorithm\cite{LOCOCV}. LOCO-CV (along with many other linear algorithms such as principal component analysis), relies on linear separability in the data. We show that, regardless of representation being used, kernels such as RBF are advantageous in reducing cluster size unevenness, and so should be strongly considered where such linear algorithms are applied. This reduction in cluster size unevenness tackles previously discussed caveats to LOCO-CV and results in more reliable model convergence (\cref{fig:LOCOCVConvergence}). 
\par We examine the use of random projections to featurise chemical compositions to be used with kernelised LOCO-CV. As for other CBFVs examined, random projections used in conjunction with kernel methods produce more even clusters than without kernel methods. However, no representation (either CBFV or random projection) consistently resulted in more even clusters than all other representations. While most of the time CBFVs found more even clusters than random projections (with the exception of \textit{CompVec}), these findings were not universal across datasets tested. Kernel methods applied to random projections resulted in cluster sizes being even enough so as to be usable in the LOCO-CV algorithm without negatively impacting conclusions drawn from measurements taken using this method. 
\par Random projections and kernelised LOCO-CV can be used together to create a generalised workflow for evaluating the extrapolatory power of a supervised machine learning algorithm, which can be used regardless of input representation to the machine learning algorithm in question. This can be combined with using a random projection as input representation to the machine learning algorithm to see a baseline measure of extrapolatory power which prospective CBFVs can be compared against to measure their usefulness.

\subsection{Conclusion}
We demonstrate random projections are a generic and powerful way to featurise compositions for material property prediction. This is motivated by fundamental principles discussed in the Johnson-Lendenstrauss lemma\cite{johnsonLindenstrauss}; randomly projecting a composition vector can be used to move such vectors into a different dimensional space while preserving relationships between points in a dataset (within some error). These random projections have only a single hyperparameter (the size of the projection), which allows us to isolate the relationships between the dimensionality of a representation, and the predictive performance of algorithms trained using that representation. Random projections can be used as a baseline representation to examine what benefit is added by domain knowledge imbued into CBFVs. 
\par We investigate how common CBFVs could be used in ten property prediction tasks from literature, in order to establish what advantage domain knowledge offers in constructing such vectors. With the notable exception of band gap prediction tasks, CBFVs engineered to incorporate domain knowledge do not substantially outperform an equal sized random projection for most prediction tasks investigated here. If the purpose of an ML model is to maximise predictive performance, the choice of using one of many complex representations (\textit{e.g}., CBFVs) should be justified by demonstrating an advantage over a random projection of the same size. 
\par We present kernelised LOCO-CV to overcome issues with imbalanced cluster sizes that often occur when performing linear clustering on material sciences datasets. The application of kernel methods, such as the RBF examined here, to data before K-means clustering leads to more even cluster sizes across many different datasets and input representations. Further, using these kernel-modified clusters in LOCO-CV led to more reliable model convergence in the models examined here. Applying kernels in LOCO-CV is independent of representations used by a supervised machine learning algorithm, so we strongly suggest that researchers looking to deploy LOCO-CV use the kernelised version presented here. Both random projections and kernelised LOCO-CV can be implemented independently or together. 
\par We trained over 70 random forest models across ten property predictions tasks found in the materials science literature to show that random projections are a reliable baseline to use when evaluating a CBFV. We have also evaluated over 36,000 K-means clustering applications, on the datasets used in these tasks as well as on the ICSD, and have shown that applying kernel functions to these data before K-means clustering results in more evenly sized clusters, and more reliable model convergence when these clusters are used in LOCO-CV. Our findings provide a basis for materials scientists in selecting and evaluating representations and laying out evaluation workflows. 
\section{Methods}
Above experiments were implemented in Python using RF, K-means clustering and kernel method algorithms from the sci-kit learn library~\cite{skl}. Hyperparameters of all sci-kit learn algorithms were set to default as of version 2.4.1, with the exception of the value of $k$ for K-means clustering which was varied between 2 and 10 as needed for the LOCO-CV algorithm. While data standardisation was sometimes done before application of K-means clustering (as detailed in the supplementary information section S2), data standardisation was not done before application use of RFs as by their nature RFs consider dimensions independently making such standardisation redundant. 
\par Graphs were plotted with the MatPlotLib library~\cite{plt} with the exception of~\cref{fig:LOCOCVConvergence} which was also uses the Seaborn library~\cite{Seaborn}. Featurisation was done using the utilities provided with the github associated with Murdock \textit{et al}.~\cite{Murdock}, with the exception of \textit{CompVec} which was implemented from scratch, and case study specific featurisations, which were obtained in supplementary information for the relevant case study. All implementations, are made available through the associated git repository as are data used in this study~\cite{myGit}.
\section{Acknowledgements}
We thank the Leverhulme Research Centre for Functional Material Design, for funding this research. Additionally we thank Taylor Sparks, Valentin Stanev, and Aron Walsh for correspondence which assisted in recreation of previous work. 
\section{Competing Interests}
The authors have no competing interests to declare.
\section{Contributions}
 SD performed initial experiments into effect of representation on RF performance and kernel method application on K-means clustering and LOCO-CV. These were experiments were then built upon by SD in discussion with VG, DB, MWG. SD wrote the first draft, and led all the redrafting. All authors input at each redrafting stage
\section{Funding}
This research was fully funded by the Leverhulme Research Centre for Functional Material Design.





\bibliography{references} 
\bibliographystyle{rsc} 

\end{document}